\documentclass[twocolumn]{ceurart}

\usepackage[utf8]{inputenc}
\usepackage[T1]{fontenc}

\usepackage{caption}

\usepackage[rgb]{xcolor}
\usepackage{tikz}
\usepackage{pgfplots}
\pgfplotsset{compat=newest}

\definecolor{blue}{RGB}{4,0,204}
\definecolor{red}{RGB}{255,123,129}

\usepackage{tipa}
\usepackage[force]{covington}

\usepgfplotslibrary{groupplots}
\begin{document}

\copyrightyear{2023}
\copyrightclause{Copyright for this paper by its authors.
  Use permitted under Creative Commons License Attribution 4.0
  International (CC BY 4.0).}

\conference{SEPLN 2023: 39\textsuperscript{th} International Conference of the Spanish Society for Natural Language Processing}

\title{ALBERTI, a Multilingual Domain Specific Language Model for Poetry Analysis}

% \author{Anonymous}[%
% orcid=0000-XXXX-XXXX-XXXX,
% email=email@email.com,
% ]

\author[1]{Javier {de la Rosa}}[%
orcid=0000-0002-9143-5573,
email=versae@nb.no,
]
\address[1]{National Library of Norway, Norway}

\author[2]{Alvaro {Perez Pozo}}[%
orcid=0000-0001-7116-9338,
email=alvaro.perez@linhd.uned.es,
]
\address[2]{Universidad Nacional de Educación a Distancia, Spain}

\author[2]{Salvador Ros}[%
orcid=0000-0001-6330-4958,
email=sros@scc.uned.es,
]

\author[3]{Elena Gonzalez-Blanco}[
orcid=0000-0002-0448-1812,
email=egonzalezblanco@faculty.ie.edu,
]
\address[3]{IE University, Spain}
% School of Human Sciences and Technology \\

% \author[4]{Manfred Jeusfeld}[%
% orcid=0000-0002-9421-8566,
% email=Manfred.Jeusfeld@acm.org,
% url=http://conceptbase.sourceforge.net/mjf/,
% ]
% \address[4]{University of Skövde, Högskolevägen 1, 541 28 Skövde, Sweden}

\begin{abstract}
The computational analysis of poetry is limited by the scarcity of tools to automatically analyze and scan poems. In a multilingual settings, the problem is exacerbated as scansion and rhyme systems only exist for individual languages, making comparative studies very challenging and time consuming. In this work, we present \textsc{Alberti}, the first multilingual pre-trained large language model for poetry. Through domain-specific pre-training (DSP), we further trained multilingual BERT on a corpus of over 12 million verses from 12 languages. We evaluated its performance on two structural poetry tasks: Spanish stanza type classification, and metrical pattern prediction for Spanish, English and German. In both cases, \textsc{Alberti} outperforms multilingual BERT and other tranformers-based models of similar sizes, and even achieves state-of-the-art results for German when compared to rule-based systems, demonstrating the feasibility and effectiveness of DSP in the poetry domain.
\end{abstract}


\maketitle

\section{Introduction}

% TODO: Describe multilingual poetry analysis

% TODO: Why is it important

% Having multilingual tools for scansion and analysis of the poetic language may enable large scale examinations across poetry traditions and uncover patterns difficult to find otherwise.

% TODO: What are the challenges

% TODO: What are we doing and how

% NUEVO:

Poetry analysis is the process of examining the elements of a poem to understand its meaning. To analyze poetry, readers must examine its words and phrasing from the perspectives of rhythm, sound, images, obvious meaning, and implied meaning. Scansion, a common approach to analyze metrical poetry, is the method or practice of determining and usually graphically representing the metrical pattern of a line of verse. It breaks down the anatomy of a poem by marking the metrical pattern of a poem by breaking each line of verse up into feet and highlighting the stressed and unstressed syllables \cite{lennard2006poetry}.

Having multilingual tools for scansion and analysis of poetic language enables large-scale examinations of poetry traditions, helping researchers identify patterns and trends that may not be apparent through an examination of a single tradition or language \cite{vsecla2022semantics}. By using multilingual tools, scholars can compare and contrast different poetic forms, structures, and devices across languages and cultures, allowing them to uncover similarities and differences and gain a more comprehensive understanding of poetic expression.

% For example, multilingual tools could help identify patterns in meter and rhyme schemes that are common across different languages. They could also help identify common themes, motifs, and symbols that are present across different poetic traditions. Additionally, multilingual tools can aid in the analysis of poetic language and form, such as meter, syntax, and diction, allowing scholars to better understand the stylistic and technical aspects of poetry.

% However, there are several challenges associated with multilingual poetry analysis. For example, working with multiple languages may require expertise in multiple linguistic and cultural traditions. Additionally, there may be issues with translation and interpretation, as poetry often relies quite heavily on figurative language and cultural references that may be difficult to translate accurately. Another challenge is the need for sophisticated computational tools to analyze and compare poetic expression across multiple languages. This requires advanced machine learning techniques, natural language processing algorithms, and other technologies that are still being developed and refined.

However, the analysis of multilingual poetry presents significant challenges that must be overcome. It demands a deep understanding of diverse linguistic and cultural traditions, as each language brings its own unique poetic conventions and nuances. Researchers and scholars need expertise in multiple languages to navigate the intricacies of each tradition accurately. Additionally, translation and interpretation pose complex obstacles in multilingual poetry analysis. Figurative language, wordplay, and cultural references deeply rooted in the specific language and culture of the poem make it challenging to convey the intended meaning, emotional impact, and artistic integrity when translating. Cultural contexts, historical references, and subtle language connotations often get lost in translation, making it difficult to fully capture the essence of the original work.

Furthermore, the development of advanced computational tools is crucial for effective analysis and comparison of poetic expression across multiple languages. This requires the application of sophisticated machine learning techniques, natural language processing algorithms, and other emerging technologies. Building models that can accurately capture the unique aesthetic qualities, rhythm, rhyme, and stylistic variations in different languages is an ongoing research endeavor that requires continuous refinement and innovation.

In this work, we investigate whether domain-specific pre-training (DSP) \cite{gu2020dsp} in a multilingual poetry setting can be leveraged to mitigate some of these issues. Specifically, we introduce \textsc{Alberti}, a multilingual encoder-only BERT-based language model suited for poetry analysis. We experimentally demonstrate that \textsc{Alberti} exhibits better performance than the base model it was built on, a multilingual BERT \cite{devlin-etal-2019-bert} which was pre-trained on the 104 languages with the largest Wikipedias. And by reformulating scansion and stanza identification as classification problems, we show that \textsc{Alberti} also outperforms its based model in these downstream tasks. Moreover, we are releasing both \textsc{Alberti} and the dataset used for further training it, which consists of over 12 million verses in multiple languages.

% Overall, multilingual poetry analysis is an important and valuable field of study that can provide new insights and perspectives on the art of poetry. However, it requires specialized knowledge and advanced computational tools to fully realize its potential, and further research is needed to overcome the challenges associated with this type of analysis.

\section{Related Work}

The transformer architecture \cite{vaswani2017attention} is now pervasive in natural language processing (NLP). In the last five years, context-aware language models have revolutionized the computational modeling of language. 

In the humanities, domain specific BERT-based models \cite{devlin-etal-2019-bert} trained with the goal of predicting masked words are starting to appear. In MacBERTh, \cite{manjavacas-arevalo-fonteyn-2021-macberth}, the authors present diachronic models for pre-1950 English literature. And a new shared task on historical models for English, French, and Dutch took place last year \cite{schweter2020triple}. While pre-training these large language models from scratch is often cost-prohibitive and extremely data demanding, adjusting them to work on other domains and tasks via transfer learning requires less data and fewer resources. For poetry, computational approaches have focused primarily on generation \cite{lau2018deep,ormazabal-etal-2022-poelm} and scansion \cite{gervas2000logic,araujo2002classificador,mcaleese2007improving,ibrahim2011toward,anttila2016phonological,agirrezabal2017comparison,desisto2020interaction,delarosa2020rantanplan}, but generally in a monolingual setting. While multilingual systems exist for metrical analysis, they internally work by having different sets of rules for each language \cite{anttila2016phonological} or by building ad-hoc neural networks \cite{agirrezabal2017comparison}. To the best of our knowledge, the only attempt at multilinguality for metrical pattern prediction was introduced in \cite{delarosa2021transformers} for English, German, and Spanish, where the authors jointly fine-tune different monolingual language models and document some cross-lingual transferability when using multilingual RoBERTa \cite{liu2019roberta}. Inspired by their good results, in this work we build a domain specific language model trained on a corpus of verses in 12 languages to explore its performance on tasks of poetic nature.

\section{Methods and Data}
% In order to build such a model, we use domain specific pre-training from the mBERT weights with the same base architecture and vocabulary. We keep the same masked language modeling (MLM) objective, and continue training for 40 epochs on a corpus of 16 million verses using a Google TPUv3 VM. We use 10,000 steps for warmup, a learning rate of 1.25e-4, weight decay of 0.01, with a batch size of 256, and a maximum sequence length of 32 to accommodate only short verses. We use 10\% of the dataset for validation, reaching a final MLM accuracy after training of 57.77\%. We evaluate the validation set on a simpler and stricter MLM accuracy score for English and Spanish and compared it to that of multilingual BERT (mBERT). In this version of MLM, which we name MLM’, we randomly masked out one whole word per verse, and tried to predict it afterwards. The accuracy was calculated as the number of true predictions over the total number of masked out words. For extrinsic evaluation, we used a stanza type classification task for Spanish, and a multilingual metrical pattern prediction task.

We leverage domain-specific pre-training techniques by fine-tuning the widely used multilingual BERT (mBERT) model with the same base architecture and vocabulary for our specific domain. We adopt the masked language modeling (MLM)~\footnote{MLM is a form of self-supervised learning that involves masking some of the words in a sentence and training the model to predict them based on the surrounding words.} objective and further train the model for 40 epochs on a large corpus consisting of 12 million verses, which were sourced from various poetry anthologies. The training was conducted on a Google TPUv3 virtual machine with a batch size of 256, a learning rate of 1.25e-4, and a weight decay of 0.01. The maximum sequence length was set to 32 since verses with up to 32 tokens using the mBERT tokenizer make up for almost 99 percent of the total. Furthermore, we used a 10,000-step warmup process, which allowed the model to learn the distribution of the corpus gradually. We are naming the resulting model \textsc{Alberti}~\footnote{An homage to Spanish poet \href{https://es.wikipedia.org/wiki/Rafael_Alberti}{Rafael Alberti}.}. After training, we evaluate the model on 10\% of the corpus held out as a validation set, achieving a final global MLM accuracy of 57.77\%.

\subsection{PULPO}
% The training of the model was done over a new corpus we built for the occasion. The Prodigious Unannotated Literary Poetry Corpus (PULPO) is a set of multilingual verses and stanzas with over 95M words. It was created to tackle the needs of scholars interested in poetry from a machine learning perspective. Although literary corpora are becoming easier to come by, it is not the case for multilingual poetic corpora. The PULPO corpus comprises almost 16 million metrical verses from 12 different languages in 3 scripts (see Table 1). We chose these languages because of the large number of poems freely available on the Internet out of copyright or with a permissive license. The poems date from the 17th-century to contemporary poetry and a number of them also have stanza separations.

% NUEVO:
The training of the model was done over a new corpus we built for the occasion. The Prolific Unannotated Literary Poetry Corpus (PULPO) is a set of multilingual verses and stanzas with over 72 million words. It was created to tackle the needs of scholars interested in poetry from a machine learning perspective. Although poetry is a fundamental aspect of human expression that has been around for millennia, the study of poetry from a machine learning perspective is still in its infancy, largely due to the scarcity of poetic corpora. And while literary corpora are becoming more readily available, multilingual poetic corpora remain elusive. The lack of such corpora presents a major challenge for researchers interested in natural language processing (NLP) and machine learning (ML) applied to poetry.

\begin{table}[!htp]
\caption{Number of deduplicated verses and their words per language in PULPO.}
\centering
\begin{tabular}{lrr}
\toprule
Language & Verses & Words \\
\midrule
English &            2,945,882 &           21,129,934 \\
Czech &            1,888,680 &           10,451,247 \\
German &            1,583,504 &            9,686,032 \\
Arabic &            1,388,461 &            6,539,196 \\
Finnish &            1,046,162 &            3,377,398 \\
Spanish &              912,951 &            5,478,627 \\
Italian &              661,526 &            4,358,541 \\
Russian &              628,719 &            3,458,928 \\
Hungarian &              495,167 &            2,444,775 \\
Chinese &              436,384 &            1,649,711 \\
Portuguese &              346,974 &            2,302,886 \\
French &              223,928 &            1,672,759 \\
\midrule
\textbf{Total} & 12,558,338 & 72,550,034 \\
\bottomrule
\end{tabular}
\label{tab.pulpo}
\end{table}
% \vspace{-1em}

The PULPO corpus comprises over 12 million deduplicated metrical verses from 12 different languages in 3 scripts (see Tables \ref{tab.pulpo} and \ref{tab.corpora}). We chose these languages because of the large number of poems freely available on the Internet out of copyright or with a permissive license. The poems date from the 15th-century to contemporary poetry and a number of them also have stanza separations. This makes the corpus a valuable resource for multilingual NLP and machine learning research. In addition, the corpus includes poems from various historical periods and literary traditions, providing a diverse range of poetic styles and forms.

\subsection{Stanzas}
To further evaluate the performance of the model, we conduct extrinsic evaluations using two different tasks. First, a stanza-type classification task for Spanish poetry. This task aims to assess the ability of the model to distinguish between different stanza types, such as tercet, quatrain, and sestina (see Table \ref{tab.stanza} for an example).

\begin{table}[h]
\centering
\begin{tabular}{lr}
\toprule
Verse & Length and Rhyme \\
\midrule
Escribí en el aren\textbf{a}l & 8a \\
los tres nombres de la vida: & 8- \\
vida, muerte, am\textbf{o}r. & 6b \\
Una ráfaga de m\textbf{a}r, & 8a \\
tantas claras veces d\textbf{a}, & 8a \\
vino y nos borr\textbf{ó}. & 6b \\
\bottomrule
\end{tabular}
\caption{Example of a stanza with its metrical length and rhyme scheme.}
\label{tab.stanza}
\end{table}

A stanza, which is considered the fundamental structural unit of a poem, serves to encapsulate themes or ideas \cite{kirszner2007literature}. Comprised of verses, stanzas are influenced by the writing styles and historical preferences of authors. The Spanish tradition boasts a rich abundance of stanza types, rendering their identification a challenging and intricate task. Generally, three factors contribute to the identification of a stanza: metrical length, rhyme type, and rhyme scheme \cite{caparros2014metrica,jauralde2020metrica,quillis2000metrica,torre2000metrica}. Consequently, the classification of stanzas can be approached in three stages \cite{caparros2014metrica}:

\begin{enumerate}
    \item Calculation of the metrical length per verse. This process typically involves counting the number of syllables while considering rhetorical devices that may alter this count (e.g., syneresis, synalephas). In some cases, the pattern formed by these verse lengths can assist in determining the stanza type.
    \item Determination of the rhyme type. When the sounds after the final stressed syllable of each verse match, it is known as consonance rhyme. Alternatively, assonance rhyme involves the matching of vowel sounds while disregarding consonant sounds. However, there are stanza types where this distinction becomes irrelevant.
    \item Extraction of the rhyme scheme. The rhyme scheme is established based on the verses that share a rhyme.
\end{enumerate}

Following \cite{perez2021bridge}, we approached stanza type identification as a classification task. We used their 5,005 Spanish stanzas containing between 12 and 170 examples for each of the 45 different types of stanzas\footnote{An extra stanza type `unknown' was ignored in this study as it does account for anything not recognized as any of the other stanza types}, and used the already existing splits of 80\% for training, 10\% for validation, and 10\% for testing.

\subsection{Scansion}
Second, a multilingual scansion task aimed at testing the ability of the model to predict the metrical pattern of a given verse in different languages. The scanning of a verse relies on assigning stress correctly to the syllables of the words. This process can be influenced by rhetorical figures and individual traditions. The synalepha is a common device in Spanish, English, and German poetry, which combines separate phonological groups into a single unit for metrical purposes. Syneresis and dieresis are two other devices that operate similarly but within the word, either joining or splitting syllables. The meter of a verse can be seen as a sequence of stressed and unstressed syllables, represented by the symbols `$+$' and `$-$', respectively. Examples \ref{example.1}, \ref{example.2}, and \ref{example.3} from \cite{delarosa2021transformers} illustrate verses with metrical lengths of 8, 10, and 7 syllables in Spanish, English, and German, respectively. These examples also demonstrate the resulting metrical pattern after applying (or breaking, as in the case for `\textit{la-her}' in the Spanish verse) synalepha, represented by `\textbottomtiebar{ }', and considering the stress of the last word as it may affect the metrical length in Spanish poetry.

\begin{example}\label{example.1}
\textit{cubra de nieve la hermosa cumbre}\footnote{{"[It]} cover with snow the beautiful summit."}\\
\textit{\textbf{cu}-bra-de-\textbf{nie}-ve-la-her-\textbf{mo}-sa-\textbf{cum}-bre} \\
$+--+---+-+-$ 11 \\
\hspace*{\fill}(Garcilaso de la Vega)\quad \\
\end{example}

% \begin{example}\label{example.1}
% \textit{Cuando el alba me despierta} \\
% \textit{Cuan-d\textbottomtiebar{oe}l-\textbf{al}-ba-me-des-\textbf{pier}-ta} \\
% $--+---+-$ 8 \\
% (Miguel de Unamuno)
% \end{example}

\begin{example}\label{example.2}
\textit{Our foes to conquer on th’ embattled plain;} \\
\textit{Our-\textbf{foes}-to-\textbf{con}-quer-on-t\textbottomtiebar{h'e}m-\textbf{bat}-tled-\textbf{pla}in;} \\
$-+-+---+-+$ 10 \\
\hspace*{\fill}(Rhys Prichard)\quad \\
\end{example}

% \end{multicols}

\begin{example}\label{example.3}
\textit{Leise lausch’ ich an der Thür}\footnote{"I quietly listen at the door"} \\
\textit{\textbf{Lei}-se-\textbf{la}-sch\textbottomtiebar{u'i}ch-\textbf{an}-der-\textbf{Thür} \\}
$+-+-+-+$ 7 \\
\hspace*{\fill}(Adolf Schults)\quad \\
\end{example}

In order to measure the performance of \textsc{Alberti}, we follow the experimental design in \cite{delarosa2021transformers} and use their chosen datasets of verses manually annotated with syllabic stress for English, German, and Spanish. For the Spanish corpus, the \textit{Corpus de Sonetos de Siglo de Oro} \cite{navarro2016metrical} was used. This TEI-XML annotated corpus consists of hendecasyllabic verses from Golden Age Spanish authors. A subset of 100 poems initially used for evaluating the ADSO Scansion system \cite{navarro2017metrical} was selected for testing, while the remaining poems were split for training and evaluation.

\begin{table} [htbp]
\caption{\label{tab.scansion}Number of verses for each language in the metrical pattern prediction datasets.}
\begin{center}
\begin{tabular} {lrrr}
  \toprule
  {\bf } & {\bf Train } & {\bf Evaluation } & {\bf Test }\\
  \midrule
  Spanish & 7,327 & 1,421 & 1,401 \\
  English & 708 & 152 & 153 \\
  German & 775 & 167 & 168 \\
\bottomrule
\end{tabular}
\end{center}
\end{table}

Unfortunately, suitable annotated corpora of comparable scale were not found for English and German. Instead, an annotated corpus of 103 poems from \textit{For Better For Verse} \cite{tucker2011poetic} was used for English, and a manually annotated corpus from \cite{haider2018supervised,haider2020po} was used for German. The German corpus contains 158 poems which cover the period from 1575 to 1936. Around 1200 lines have been annotated in terms of syllable stress, foot boundaries, caesuras and line main accent. These corpora were divided into train, evaluation, and test sets, following a 70-15-15 split. Table \ref{tab.scansion} shows the number of verses per language and split .
% In this study, we use the binary codification for clarity, with 0 indicating an unstressed syllable and 1 a stressed one.

\section{Evaluation and Results}

% Probability score function
% In [17]: def prob(sentence, model, tokenizer):
%     ...:     model_inputs = tokenizer(sentence, add_special_tokens=False, return_tensors="pt")
%     ...:     scores = []   
%     ...:     count = 0    
%     ...:     for input_index in range(len(model_inputs["input_ids"][0])):
%     ...:         masked_token = tokenizer.decode(model_inputs["input_ids"][0][input_index], skip_special_tokens=True)
%     ...:         if len(masked_token) > 0:
%     ...:             model_inputs["input_ids"][0][input_index] = tokenizer.mask_token_id
%     ...:             scores.append(fill(model_inputs, targets=[masked_token])[0]["score"])
%     ...:             model_inputs = tokenizer(sentence, add_special_tokens=False, return_tensors="pt")
%     ...:                                              
%     ...:     return math.pow(math.prod(scores), -1 / len(scores))

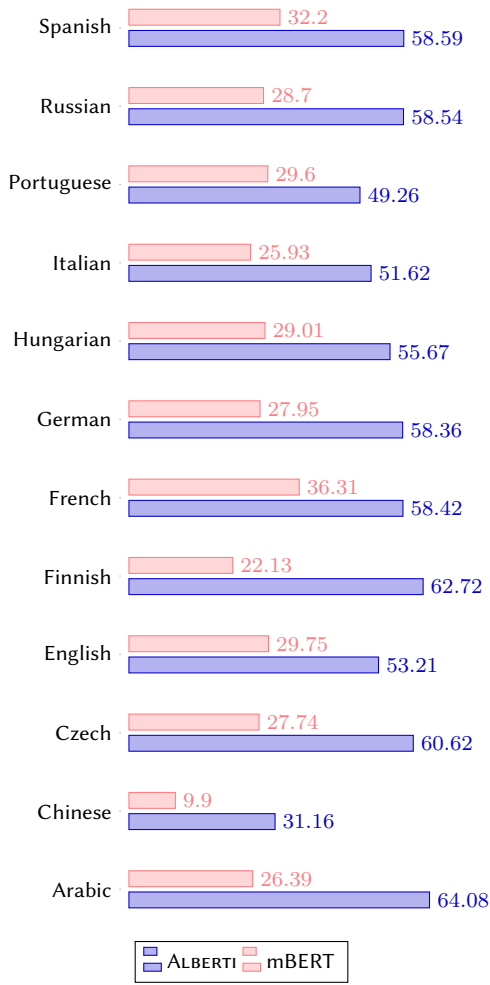
\begin{figure}[!h]
    \centering
    \begin{tikzpicture}
    \begin{axis}[
        xbar,
        y axis line style = { opacity = 0 },
        axis x line       = none,
        tickwidth         = 0pt,
        enlarge y limits  = 0.2,
        enlarge x limits  = 0.02,
        width=8cm,
        height=500pt,
        xlabel={Accuracy},
        xmin=0,
        xmax=100,
        ytick=data,
        yticklabels={Arabic, Chinese, Czech, English, Finnish, French, German, Hungarian, Italian, Portuguese, Russian, Spanish},
        yticklabel style={text width=2cm,align=right},
        legend style={at={(0.25,0.1)},anchor=north,legend columns=-1},
        bar width=6pt,
        nodes near coords,
        nodes near coords align={horizontal},
        nodes near coords style={font=\small},
    ]
    \addplot coordinates {(64.0829,0) (31.1552,1) (60.6241,2) (53.2082,3) (62.7208,4) (58.4215,5) (58.3583,6) (55.6652,7) (51.6227,8) (49.2606,9) (58.5366,10) (58.5948,11)};
    \addplot coordinates {(26.3913,0) (09.8953,1) (27.7384,2) (29.7501,3) (22.1314,4) (36.3056,5) (27.9479,6) (29.0083,7) (25.9347,8) (29.6045,9) (28.6993,10) (32.2025,11)};
    \legend{\textsc{Alberti}, mBERT}
    \end{axis}
    \end{tikzpicture}
    \caption{Masked Language Modeling accuracy (\%) on the validation set of PULPO for \textsc{Alberti} (blue) and mBERT (red). Higher is better.}
    \label{fig:mlm}
\end{figure}

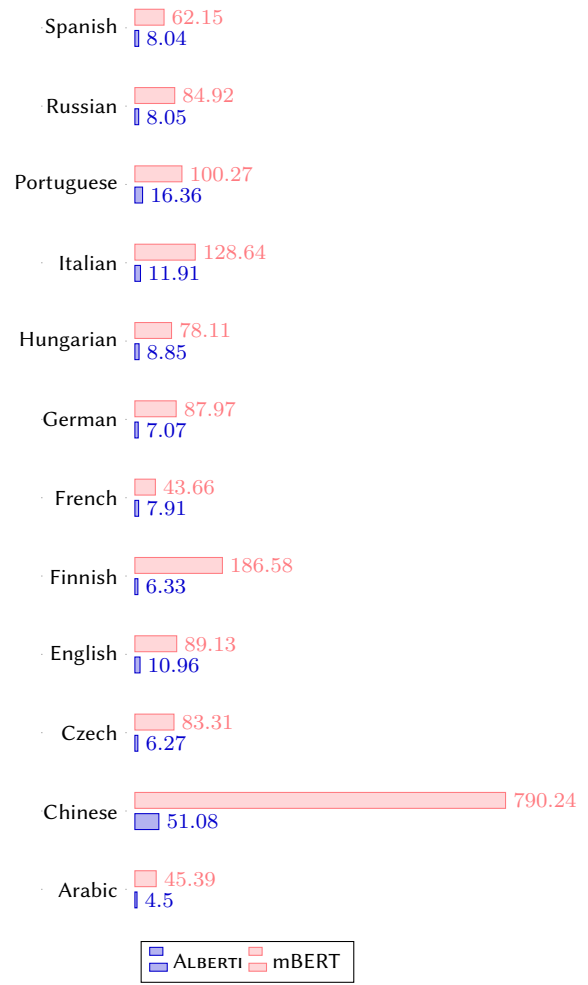
\begin{figure}[!h]
    \begin{tikzpicture}
    \begin{axis}[
        xbar,
        y axis line style = { opacity = 0 },
        axis x line       = none,
        tickwidth         = 0pt,
        enlarge y limits  = 0.2,
        enlarge x limits  = 0.02,
        width=8cm,
        height=500pt,
        xlabel={Perplexity},
        xmin=0,
        xmax=1000,
        ytick=data,
        yticklabels={Arabic, Chinese, Czech, English, Finnish, French, German, Hungarian, Italian, Portuguese, Russian, Spanish},
        yticklabel style={text width=2cm,align=right},
        legend style={at={(0.25,0.1)},anchor=north,legend columns=-1},
        bar width=6pt,
        nodes near coords,
        nodes near coords align={horizontal},
        nodes near coords style={font=\small},
    ]
    \addplot coordinates {(4.504221,0) (51.075327,1) (6.270954,2) (10.957538,3) (6.330115,4) (7.909655,5) (7.070457,6) (8.849815,7) (11.911083,8) (16.362507,9) (8.054538,10) (8.041797,11)};
    \addplot coordinates {(45.387927,0) (790.236746,1) (83.308096,2) (89.133680,3) (186.577666,4) (43.663215,5) (87.971807,6) (78.108549,7) (128.643954,8) (100.267265,9) (84.915548,10) (62.150973,11)};
    \legend{\textsc{Alberti}, mBERT}
    \end{axis}
    \end{tikzpicture}
    \caption{Perplexity proxy scores on the validation set of PULPO for \textsc{Alberti} (blue) and mBERT (red). Lower is better.}
    \label{fig:ppl}
\end{figure}

After training, we evaluated the resulting model \textsc{Alberti} on several fronts. For intrinsic evaluation, we used the aforementioned MLM metric as well as a perplexity proxy score based on the predicted token probabilities. We calculated these metrics for every language on the validation set of PULPO for both \textsc{Alberti} and mBERT. As shown in Figure~\ref{fig:mlm}, the MLM accuracy of \textsc{Alberti} is generally higher than that of mBERT for all languages. The gains of \textsc{Alberti} against mBERT range from +19.65 percentage points for Portuguese to +40.59 for Finnish. A similar trend is shown for our perplexity proxy score in Figure~\ref{fig:ppl}, with clear gains of \textsc{Alberti} over mBERT across the board, ranging from -35.75 for French to staggering -739.16 points for Chinese. The stark difference for Chinese could be a result of differences in the way text is represented in that language in both the pre-training corpus of mBERT and PULPO.

For extrinsic evaluation, we also evaluated \textsc{Alberti} against mBERT for stanza classification and metrical pattern prediction. We chose the best performing models on the validation set over a small grid search of learning rates  $10^{-5}$, $3 \times 10^{-5}$, and $5 \times 10^{-5}$, for 3, 5, and 10 epochs, and warmup of 0 and 10\% of the steps. Figure \ref{fig:roc} shows the ROC curves of each stanza type versus the rest for both \textsc{Alberti} and mBERT, with higher areas under the curve (AUC) in 29 out of the 45 stanza types for \textsc{Alberti}, and in 16 out of 45 for mBERT. Table \ref{tab.stanzas} shows F1 and accuracy macro scores for each model, with \textsc{Alberti} outperforming mBERT by a small percentage. Interestingly, our baseline fine-tuned mBERT model scores better than the monolingual Spanish BETO \cite{caneteCFP2020} reported in \cite{perez2021bridge}. Nonetheless, the combination of the rule-based system Rantanplan \cite{delarosa2020rantanplan} with an expert system remains state of the art for stanza classification.

\begin{table}[h]
\caption{F1 scores on stanza classification. Best neural model scores in \textbf{bold}. Rule-based systems \textit{italicized}.}
\centering
\begin{tabular}{lrr}
\toprule
Model & F1 & Accuracy \\
\midrule
mBERT & 57.51 & 61.94 \\
\textsc{Alberti} & \textbf{59.33} & \textbf{63.64} \\
\midrule
BETO \cite{perez2021bridge} & -- & 42.12 \\
\textit{Rantanplan \cite{delarosa2020rantanplan} + Expert System \cite{perez2021bridge}} & -- & \textit{88.51} \\
\bottomrule
\end{tabular}
\label{tab.stanzas}
\end{table}

The prediction of metre was approached as a multi-class binary classification task, i.e., one class per syllable where each syllable can be stressed (strong) or unstressed (weak). After a grid search with roughly the same hyperparameters as in \cite{delarosa2021transformers}, \textsc{Alberti} outperforms mBERT for every language, as shown in Table~\ref{tab.metrical}. When compared to other similarly sized models (English RoBERTa \cite{liu2019roberta} and multilingual XLM RoBERTa \cite{conneau2019xlm-roberta}) as reported in \cite{delarosa2021transformers}, it still performs better for English and German. Lastly, \textsc{Alberti} achieves a new state-of-the-art for German, as it performs better than both the large version of XLM RoBERTa and the rule-based system Metricalizer \cite{bobenhausen2011metricalizer2}.

\begin{table}[h]
\caption{Accuracy on metrical pattern prediction. Best neural model scores in \textbf{bold}. Rule-based systems \textit{italicized}.}
\centering
\begin{tabular}{lrrr}
\toprule
Model & Spanish & English &  German \\
\midrule
mBERT & 88.15 & 35.71 & 39.52 \\
\textsc{Alberti} & 91.15 & \textbf{49.34} &  \textbf{56.29} \\
\midrule
RoBERTa (base) \cite{delarosa2021transformers} & 87.37 & 36.21 & 43.11  \\
XLM RoBERTa (base) \cite{delarosa2021transformers} & \textbf{92.15} & 40.79 & 46.11  \\
% XLM RoBERTa (large) & 93.29 & 50.66 & 48.50 \\
\textit{Rantanplan \cite{delarosa2020rantanplan}} & \textit{96.23} & -- & -- \\
\textit{Poesy \cite{algee2014stanford}} & -- & \textit{38.16}  & -- \\
\textit{Metricalizer \cite{bobenhausen2011metricalizer2}} & -- & -- & \textit{44.91} \\
\bottomrule
\end{tabular}
\label{tab.metrical}
\end{table}

\usetikzlibrary{backgrounds,calc}
\newcommand\myaxiswidth{2.0cm}

\begin{figure*}
    \centering

\begin{tikzpicture}
  \begin{groupplot}[
        clip=false,
        axis x line*=bottom,
        axis y line*=left,
        ylabel=TPR,
        xlabel=FPR,
        group style={
            group size=6 by 8,
            x descriptions at=edge bottom,
            y descriptions at=edge left,
            horizontal sep=10pt,
            vertical sep=20pt},
         ymin=0.0,ymax=1.0,
         xmin=0.0,xmax=1.0,
         width=\myaxiswidth,
         height=\myaxiswidth,
         scale only axis,
         enlargelimits=0.05,
         title style={
            text centered,
            yshift=-4.5pt,
            fill=black!10,
            minimum height=1pt,
            minimum width=\myaxiswidth},
         % domain=0:45, %just for example
         legend style={draw=none, fill=none},
         legend pos=south east,
         ]

    \nextgroupplot[title=Cantar]
        \addlegendimage{empty legend}
        \addplot[blue] coordinates {(0.0,0.0) (0.0,0.034) (0.0,0.069) (0.005,0.069) (0.005,0.103) (0.006,0.103) (0.006,0.138) (0.007,0.138) (0.007,0.172) (0.008,0.172) (0.008,0.207) (0.009,0.207) (0.009,0.241) (0.01,0.241) (0.01,0.276) (0.012,0.276) (0.012,0.31) (0.022,0.31) (0.022,0.345) (0.025,0.345) (0.025,0.414) (0.03,0.414) (0.03,0.448) (0.031,0.448) (0.031,0.517) (0.036,0.517) (0.036,0.586) (0.039,0.586) (0.039,0.621) (0.04,0.621) (0.04,0.655) (0.05,0.655) (0.05,0.69) (0.065,0.69) (0.065,0.724) (0.074,0.724) (0.074,0.759) (0.11,0.759) (0.11,0.828) (0.113,0.828) (0.113,0.862) (0.124,0.862) (0.124,0.897) (0.126,0.897) (0.126,0.931) (0.149,0.931) (0.149,0.966) (0.286,0.966) (0.286,1.0) (0.571,1.0) (0.573,1.0) (1.0,1.0)};
        \addplot[red] coordinates {(0.0,0.0) (0.0,0.034) (0.003,0.034) (0.003,0.069) (0.005,0.069) (0.005,0.103) (0.006,0.103) (0.006,0.241) (0.007,0.241) (0.007,0.345) (0.01,0.345) (0.01,0.379) (0.027,0.379) (0.027,0.414) (0.029,0.414) (0.029,0.483) (0.037,0.483) (0.037,0.517) (0.04,0.517) (0.04,0.552) (0.042,0.552) (0.042,0.586) (0.043,0.586) (0.043,0.69) (0.061,0.69) (0.061,0.724) (0.062,0.724) (0.062,0.759) (0.079,0.759) (0.079,0.793) (0.08,0.793) (0.08,0.828) (0.082,0.828) (0.082,0.862) (0.084,0.862) (0.084,0.897) (0.095,0.897) (0.095,0.931) (0.13,0.931) (0.13,0.966) (0.17,0.966) (0.17,1.0) (0.647,1.0) (0.649,1.0) (1.0,1.0)};
        \addlegendentry{\textbf{AUC}}
        \addlegendentry{0.9456}
        \addlegendentry{\textbf{0.9573}}

    \nextgroupplot[title=Chamberga]
        \addlegendimage{empty legend}
        \addplot[blue] coordinates {(0.0,0.0) (0.0,0.091) (0.0,0.545) (0.002,0.545) (0.002,0.636) (0.005,0.636) (0.005,0.818) (0.014,0.818) (0.014,0.909) (0.064,0.909) (0.064,1.0) (0.943,1.0) (0.945,1.0) (1.0,1.0)};
        \addplot[red] coordinates {(0.0,0.0) (0.0,0.091) (0.0,0.273) (0.001,0.273) (0.001,0.455) (0.006,0.455) (0.006,0.545) (0.01,0.545) (0.01,0.636) (0.013,0.636) (0.013,0.727) (0.015,0.727) (0.015,0.818) (0.017,0.818) (0.017,0.909) (0.022,0.909) (0.022,1.0) (0.956,1.0) (0.958,1.0) (1.0,1.0)};
        \addlegendentry{\textbf{AUC}}
        \addlegendentry{0.9918}
        \addlegendentry{\textbf{0.9923}}

    \nextgroupplot[title=Copla arte mayor]
        \addlegendimage{empty legend}
        \addplot[blue] coordinates {(0.0,0.0) (0.0,0.071) (0.0,0.429) (0.001,0.429) (0.001,0.857) (0.008,0.857) (0.008,0.929) (0.017,0.929) (0.017,1.0) (0.554,1.0) (0.556,1.0) (1.0,1.0)};
        \addplot[red] coordinates {(0.0,0.0) (0.0,0.071) (0.0,0.214) (0.001,0.214) (0.001,0.643) (0.002,0.643) (0.002,0.786) (0.003,0.786) (0.003,0.857) (0.004,0.857) (0.004,0.929) (0.007,0.929) (0.007,1.0) (0.817,1.0) (0.819,1.0) (1.0,1.0)};
        \addlegendentry{\textbf{AUC}}
        \addlegendentry{0.9978}
        \addlegendentry{\textbf{0.9983}}

    \nextgroupplot[title=Copla arte menor]
        \addlegendimage{empty legend}
        \addplot[blue] coordinates {(0.0,0.0) (0.0,0.071) (0.0,0.214) (0.001,0.214) (0.001,0.5) (0.002,0.5) (0.002,0.786) (0.003,0.786) (0.003,0.857) (0.006,0.857) (0.006,0.929) (0.037,0.929) (0.037,1.0) (0.719,1.0) (0.721,1.0) (1.0,1.0)};
        \addplot[red] coordinates {(0.0,0.0) (0.0,0.071) (0.002,0.071) (0.002,0.286) (0.003,0.286) (0.003,0.357) (0.004,0.357) (0.004,0.429) (0.005,0.429) (0.005,0.5) (0.007,0.5) (0.007,0.571) (0.01,0.571) (0.01,0.643) (0.011,0.643) (0.011,0.714) (0.02,0.714) (0.02,0.786) (0.027,0.786) (0.027,0.929) (0.028,0.929) (0.028,1.0) (0.871,1.0) (0.873,1.0) (1.0,1.0)};
        \addlegendentry{\textbf{AUC}}
        \addlegendentry{\textbf{0.9959}}
        \addlegendentry{0.9894}

    \nextgroupplot[title=Copla castellana]
        \addlegendimage{empty legend}
        \addplot[blue] coordinates {(0.0,0.0) (0.0,0.036) (0.0,0.214) (0.001,0.214) (0.001,0.357) (0.002,0.357) (0.002,0.429) (0.004,0.429) (0.004,0.536) (0.007,0.536) (0.007,0.571) (0.009,0.571) (0.009,0.607) (0.01,0.607) (0.01,0.75) (0.011,0.75) (0.011,0.786) (0.012,0.786) (0.012,0.821) (0.015,0.821) (0.015,0.857) (0.022,0.857) (0.022,0.893) (0.026,0.893) (0.026,0.929) (0.039,0.929) (0.039,0.964) (0.079,0.964) (0.079,1.0) (0.888,1.0) (0.89,1.0) (1.0,1.0)};
        \addplot[red] coordinates {(0.0,0.0) (0.0,0.036) (0.0,0.214) (0.001,0.214) (0.001,0.393) (0.004,0.393) (0.004,0.429) (0.007,0.429) (0.007,0.464) (0.008,0.464) (0.008,0.5) (0.01,0.5) (0.01,0.536) (0.012,0.536) (0.012,0.571) (0.014,0.571) (0.014,0.607) (0.016,0.607) (0.016,0.643) (0.017,0.643) (0.017,0.679) (0.03,0.679) (0.03,0.714) (0.032,0.714) (0.032,0.75) (0.047,0.75) (0.047,0.786) (0.051,0.786) (0.051,0.821) (0.071,0.821) (0.071,0.857) (0.162,0.857) (0.162,0.893) (0.226,0.893) (0.226,0.929) (0.307,0.929) (0.307,0.964) (0.475,0.964) (0.475,1.0) (0.847,1.0) (0.849,1.0) (1.0,1.0)};
        \addlegendentry{\textbf{AUC}}
        \addlegendentry{\textbf{0.9900}}
        \addlegendentry{0.9466}

    \nextgroupplot[title=Copla mixta]
        \addlegendimage{empty legend}
        \addplot[blue] coordinates {(0.0,0.0) (0.001,0.0) (0.001,0.5) (0.004,0.5) (0.004,1.0) (0.642,1.0) (0.644,1.0) (1.0,1.0)};
        \addplot[red] coordinates {(0.0,0.0) (0.0,0.5) (0.001,0.5) (0.001,1.0) (0.734,1.0) (0.736,1.0) (1.0,1.0)};
        \addlegendentry{\textbf{AUC}}
        \addlegendentry{0.9975}
        \addlegendentry{\textbf{0.9995}}

    \nextgroupplot[title=Copla real]
        \addlegendimage{empty legend}
        \addplot[blue] coordinates {(0.0,0.0) (0.0,0.036) (0.0,0.179) (0.001,0.179) (0.001,0.25) (0.002,0.25) (0.002,0.5) (0.004,0.5) (0.004,0.536) (0.005,0.536) (0.005,0.607) (0.006,0.607) (0.006,0.643) (0.007,0.643) (0.007,0.75) (0.009,0.75) (0.009,0.786) (0.01,0.786) (0.01,0.821) (0.013,0.821) (0.013,0.929) (0.014,0.929) (0.014,1.0) (0.988,1.0) (0.99,1.0) (1.0,1.0)};
        \addplot[red] coordinates {(0.0,0.0) (0.0,0.036) (0.0,0.143) (0.001,0.143) (0.001,0.179) (0.002,0.179) (0.002,0.25) (0.003,0.25) (0.003,0.286) (0.004,0.286) (0.004,0.429) (0.006,0.429) (0.006,0.643) (0.007,0.643) (0.007,0.679) (0.008,0.679) (0.008,0.714) (0.009,0.714) (0.009,0.75) (0.01,0.75) (0.01,0.821) (0.013,0.821) (0.013,0.893) (0.032,0.893) (0.032,0.964) (0.04,0.964) (0.04,1.0) (0.672,1.0) (0.674,1.0) (1.0,1.0)};
        \addlegendentry{\textbf{AUC}}
        \addlegendentry{\textbf{0.9949}}
        \addlegendentry{0.9916}

    \nextgroupplot[title=Couplet]
        \addlegendimage{empty legend}
        \addplot[blue] coordinates {(0.0,0.0) (0.0,0.036) (0.0,0.107) (0.001,0.107) (0.001,0.143) (0.003,0.143) (0.003,0.179) (0.007,0.179) (0.007,0.214) (0.009,0.214) (0.009,0.286) (0.01,0.286) (0.01,0.357) (0.012,0.357) (0.012,0.393) (0.013,0.393) (0.013,0.429) (0.014,0.429) (0.014,0.607) (0.015,0.607) (0.015,0.643) (0.017,0.643) (0.017,0.679) (0.02,0.679) (0.02,0.714) (0.022,0.714) (0.022,0.75) (0.036,0.75) (0.036,0.786) (0.04,0.786) (0.04,0.821) (0.055,0.821) (0.055,0.893) (0.096,0.893) (0.096,0.929) (0.119,0.929) (0.119,0.964) (0.162,0.964) (0.162,1.0) (0.86,1.0) (0.862,1.0) (1.0,1.0)};
        \addplot[red] coordinates {(0.0,0.0) (0.0,0.036) (0.0,0.071) (0.001,0.071) (0.001,0.25) (0.002,0.25) (0.002,0.321) (0.006,0.321) (0.006,0.393) (0.007,0.393) (0.007,0.5) (0.01,0.5) (0.01,0.607) (0.012,0.607) (0.012,0.643) (0.015,0.643) (0.015,0.679) (0.038,0.679) (0.038,0.714) (0.04,0.714) (0.04,0.75) (0.041,0.75) (0.041,0.786) (0.044,0.786) (0.044,0.821) (0.055,0.821) (0.055,0.857) (0.057,0.857) (0.057,0.893) (0.063,0.893) (0.063,0.929) (0.127,0.929) (0.127,0.964) (0.182,0.964) (0.182,1.0) (0.336,1.0) (0.338,1.0) (1.0,1.0)};
        \addlegendentry{\textbf{AUC}}
        \addlegendentry{0.9721}
        \addlegendentry{\textbf{0.9734}}

    \nextgroupplot[title=Cuaderna vía]
        \addlegendimage{empty legend}
        \addplot[blue] coordinates {(0.0,0.0) (0.0,0.037) (0.0,0.63) (0.002,0.63) (0.002,0.704) (0.003,0.704) (0.003,0.852) (0.005,0.852) (0.005,0.926) (0.007,0.926) (0.007,0.963) (0.039,0.963) (0.039,1.0) (0.592,1.0) (0.594,1.0) (0.821,1.0) (0.823,1.0) (1.0,1.0)};
        \addplot[red] coordinates {(0.0,0.0) (0.0,0.037) (0.0,0.481) (0.002,0.481) (0.002,0.778) (0.003,0.778) (0.003,0.889) (0.004,0.889) (0.004,0.926) (0.005,0.926) (0.005,0.963) (0.42,0.963) (0.42,1.0) (0.697,1.0) (0.699,1.0) (1.0,1.0)};
        \addlegendentry{\textbf{AUC}}
        \addlegendentry{\textbf{0.9973}}
        \addlegendentry{0.9832}

    \nextgroupplot[title=Cuarteta]
        \addlegendimage{empty legend}
        \addplot[blue] coordinates {(0.0,0.0) (0.001,0.0) (0.001,0.115) (0.007,0.115) (0.007,0.154) (0.013,0.154) (0.013,0.192) (0.017,0.192) (0.017,0.308) (0.018,0.308) (0.018,0.346) (0.019,0.346) (0.019,0.423) (0.023,0.423) (0.023,0.5) (0.027,0.5) (0.027,0.538) (0.032,0.538) (0.032,0.577) (0.033,0.577) (0.033,0.615) (0.037,0.615) (0.037,0.654) (0.04,0.654) (0.04,0.692) (0.049,0.692) (0.049,0.731) (0.056,0.731) (0.056,0.769) (0.057,0.769) (0.057,0.808) (0.059,0.808) (0.059,0.846) (0.071,0.846) (0.071,0.885) (0.088,0.885) (0.088,0.923) (0.118,0.923) (0.118,0.962) (0.178,0.962) (0.181,0.962) (0.253,0.962) (0.253,1.0) (1.0,1.0)};
        \addplot[red] coordinates {(0.0,0.0) (0.0,0.038) (0.001,0.038) (0.001,0.077) (0.004,0.077) (0.004,0.115) (0.006,0.115) (0.006,0.154) (0.01,0.154) (0.01,0.192) (0.011,0.192) (0.011,0.231) (0.013,0.231) (0.013,0.269) (0.014,0.269) (0.014,0.308) (0.015,0.308) (0.015,0.385) (0.017,0.385) (0.017,0.462) (0.018,0.462) (0.018,0.538) (0.019,0.538) (0.019,0.577) (0.021,0.577) (0.021,0.615) (0.028,0.615) (0.028,0.654) (0.037,0.654) (0.037,0.731) (0.038,0.731) (0.038,0.769) (0.039,0.769) (0.039,0.808) (0.041,0.808) (0.041,0.846) (0.05,0.846) (0.05,0.885) (0.056,0.885) (0.056,0.923) (0.117,0.923) (0.119,0.923) (0.137,0.923) (0.137,0.962) (0.179,0.962) (0.179,1.0) (1.0,1.0)};
        \addlegendentry{\textbf{AUC}}
        \addlegendentry{0.9579}
        \addlegendentry{\textbf{0.9677}}

    \nextgroupplot[title=Cuarteto]
        \addlegendimage{empty legend}
        \addplot[blue] coordinates {(0.0,0.0) (0.001,0.0) (0.001,0.038) (0.002,0.038) (0.002,0.192) (0.003,0.192) (0.003,0.269) (0.004,0.269) (0.004,0.308) (0.005,0.308) (0.005,0.346) (0.006,0.346) (0.006,0.385) (0.007,0.385) (0.007,0.462) (0.008,0.462) (0.008,0.5) (0.01,0.5) (0.01,0.538) (0.014,0.538) (0.014,0.577) (0.021,0.577) (0.021,0.615) (0.024,0.615) (0.024,0.692) (0.031,0.692) (0.031,0.731) (0.033,0.731) (0.033,0.769) (0.035,0.769) (0.035,0.808) (0.036,0.808) (0.036,0.846) (0.048,0.846) (0.048,0.885) (0.05,0.885) (0.05,0.923) (0.056,0.923) (0.056,0.962) (0.172,0.962) (0.172,1.0) (0.457,1.0) (0.459,1.0) (1.0,1.0)};
        \addplot[red] coordinates {(0.0,0.0) (0.0,0.038) (0.001,0.038) (0.001,0.115) (0.002,0.115) (0.002,0.154) (0.003,0.154) (0.003,0.192) (0.004,0.192) (0.004,0.231) (0.005,0.231) (0.005,0.269) (0.008,0.269) (0.008,0.346) (0.011,0.346) (0.011,0.385) (0.013,0.385) (0.013,0.423) (0.014,0.423) (0.014,0.5) (0.024,0.5) (0.024,0.538) (0.025,0.538) (0.025,0.654) (0.026,0.654) (0.026,0.692) (0.031,0.692) (0.031,0.731) (0.038,0.731) (0.038,0.769) (0.046,0.769) (0.046,0.808) (0.049,0.808) (0.049,0.846) (0.054,0.846) (0.054,0.885) (0.105,0.885) (0.105,0.923) (0.112,0.923) (0.114,0.923) (0.161,0.923) (0.161,0.962) (0.225,0.962) (0.225,1.0) (1.0,1.0)};
        \addlegendentry{\textbf{AUC}}
        \addlegendentry{\textbf{0.9767}}
        \addlegendentry{0.9647}

    \nextgroupplot[title=Cuarteto lira]
        \addlegendimage{empty legend}
        \addplot[blue] coordinates {(0.0,0.0) (0.001,0.0) (0.012,0.0) (0.012,0.167) (0.013,0.167) (0.013,0.25) (0.016,0.25) (0.016,0.417) (0.019,0.417) (0.019,0.5) (0.023,0.5) (0.023,0.583) (0.026,0.583) (0.026,0.667) (0.027,0.667) (0.028,0.75) (0.028,0.833) (0.071,0.833) (0.071,0.917) (0.086,0.917) (0.086,1.0) (1.0,1.0)};
        \addplot[red] coordinates {(0.0,0.0) (0.001,0.0) (0.008,0.0) (0.008,0.083) (0.016,0.083) (0.016,0.167) (0.017,0.167) (0.017,0.417) (0.019,0.417) (0.019,0.5) (0.02,0.5) (0.02,0.667) (0.023,0.667) (0.023,0.75) (0.027,0.75) (0.027,0.833) (0.04,0.833) (0.041,0.917) (0.604,0.917) (0.604,1.0) (1.0,1.0)};
        \addlegendentry{\textbf{AUC}}
        \addlegendentry{\textbf{0.9709}}
        \addlegendentry{0.9311}

    \nextgroupplot[title=Décima antigua]
        \addlegendimage{empty legend}
        \addplot[blue] coordinates {(0.0,0.0) (0.001,0.0) (0.001,0.267) (0.003,0.267) (0.003,0.4) (0.004,0.4) (0.004,0.467) (0.005,0.467) (0.005,0.6) (0.006,0.6) (0.006,0.667) (0.012,0.667) (0.012,0.733) (0.017,0.733) (0.017,0.867) (0.025,0.867) (0.025,0.933) (0.633,0.933) (0.635,0.933) (0.89,0.933) (0.89,1.0) (1.0,1.0)};
        \addplot[red] coordinates {(0.0,0.0) (0.0,0.067) (0.0,0.267) (0.001,0.267) (0.001,0.333) (0.002,0.333) (0.002,0.467) (0.003,0.467) (0.003,0.533) (0.005,0.533) (0.005,0.667) (0.008,0.667) (0.008,0.8) (0.009,0.8) (0.009,0.867) (0.011,0.867) (0.011,0.933) (0.112,0.933) (0.112,1.0) (0.986,1.0) (0.988,1.0) (1.0,1.0)};
        \addlegendentry{\textbf{AUC}}
        \addlegendentry{0.9336}
        \addlegendentry{\textbf{0.9889}}

    \nextgroupplot[title=Endecha real]
        \addlegendimage{empty legend}
        \addplot[blue] coordinates {(0.0,0.0) (0.0,0.071) (0.0,0.214) (0.002,0.214) (0.002,0.286) (0.005,0.286) (0.005,0.571) (0.013,0.571) (0.013,0.643) (0.017,0.643) (0.017,0.714) (0.033,0.714) (0.033,0.786) (0.042,0.786) (0.042,0.857) (0.095,0.857) (0.095,0.929) (0.131,0.929) (0.131,1.0) (0.698,1.0) (0.7,1.0) (1.0,1.0)};
        \addplot[red] coordinates {(0.0,0.0) (0.0,0.071) (0.0,0.143) (0.001,0.143) (0.001,0.214) (0.003,0.214) (0.003,0.286) (0.005,0.286) (0.005,0.357) (0.014,0.357) (0.014,0.429) (0.016,0.429) (0.016,0.5) (0.023,0.5) (0.023,0.643) (0.036,0.643) (0.036,0.786) (0.064,0.786) (0.064,0.857) (0.177,0.857) (0.177,0.929) (0.25,0.929) (0.25,1.0) (0.905,1.0) (0.907,1.0) (1.0,1.0)};
        \addlegendentry{\textbf{AUC}}
        \addlegendentry{\textbf{0.9748}}
        \addlegendentry{0.9537}

    \nextgroupplot[title=Espinela]
        \addlegendimage{empty legend}
        \addplot[blue] coordinates {(0.0,0.0) (0.0,0.037) (0.0,0.074) (0.001,0.074) (0.001,0.37) (0.003,0.37) (0.003,0.444) (0.007,0.444) (0.007,0.481) (0.008,0.481) (0.008,0.556) (0.009,0.556) (0.009,0.593) (0.011,0.593) (0.011,0.63) (0.012,0.63) (0.012,0.704) (0.018,0.704) (0.018,0.778) (0.028,0.778) (0.028,0.852) (0.049,0.852) (0.049,0.889) (0.054,0.889) (0.054,0.926) (0.062,0.926) (0.062,0.963) (0.214,0.963) (0.214,1.0) (0.616,1.0) (0.618,1.0) (1.0,1.0)};
        \addplot[red] coordinates {(0.0,0.0) (0.0,0.037) (0.001,0.037) (0.001,0.333) (0.003,0.333) (0.003,0.37) (0.005,0.37) (0.005,0.593) (0.006,0.593) (0.006,0.63) (0.007,0.63) (0.007,0.667) (0.008,0.667) (0.008,0.704) (0.009,0.704) (0.009,0.741) (0.01,0.741) (0.01,0.778) (0.026,0.778) (0.026,0.815) (0.032,0.815) (0.032,0.852) (0.038,0.852) (0.038,0.889) (0.052,0.889) (0.052,0.926) (0.06,0.926) (0.06,0.963) (0.125,0.963) (0.125,1.0) (0.443,1.0) (0.445,1.0) (1.0,1.0)};
        \addlegendentry{\textbf{AUC}}
        \addlegendentry{0.9796}
        \addlegendentry{\textbf{0.9847}}

    \nextgroupplot[title=Estr. Fco. de la Torre]
        \addlegendimage{empty legend}
        \addplot[blue] coordinates {(0.0,0.0) (0.0,0.083) (0.0,0.25) (0.001,0.25) (0.001,0.5) (0.003,0.5) (0.003,0.583) (0.015,0.583) (0.015,0.667) (0.016,0.667) (0.016,0.75) (0.017,0.75) (0.017,0.833) (0.032,0.833) (0.032,0.917) (0.099,0.917) (0.099,1.0) (0.491,1.0) (0.493,1.0) (1.0,1.0)};
        \addplot[red] coordinates {(0.0,0.0) (0.0,0.083) (0.002,0.083) (0.002,0.25) (0.003,0.25) (0.003,0.5) (0.004,0.5) (0.004,0.583) (0.014,0.583) (0.014,0.667) (0.017,0.667) (0.017,0.75) (0.021,0.75) (0.021,0.833) (0.052,0.833) (0.052,0.917) (0.223,0.917) (0.225,0.917) (0.282,0.917) (0.282,1.0) (1.0,1.0)};
        \addlegendentry{\textbf{AUC}}
        \addlegendentry{\textbf{0.9846}}
        \addlegendentry{0.9665}

    \nextgroupplot[title=Estr. manriqueña]
        \addlegendimage{empty legend}
        \addplot[blue] coordinates {(0.0,0.0) (0.0,0.036) (0.0,0.464) (0.001,0.464) (0.001,0.536) (0.002,0.536) (0.002,0.571) (0.007,0.571) (0.007,0.607) (0.015,0.607) (0.015,0.643) (0.016,0.643) (0.016,0.679) (0.017,0.679) (0.017,0.714) (0.021,0.714) (0.021,0.75) (0.027,0.75) (0.027,0.786) (0.035,0.786) (0.035,0.857) (0.043,0.857) (0.043,0.893) (0.054,0.893) (0.054,0.929) (0.087,0.929) (0.087,0.964) (0.158,0.964) (0.158,1.0) (0.28,1.0) (0.282,1.0) (1.0,1.0)};
        \addplot[red] coordinates {(0.0,0.0) (0.0,0.036) (0.0,0.321) (0.003,0.321) (0.003,0.357) (0.004,0.357) (0.004,0.464) (0.01,0.464) (0.01,0.536) (0.012,0.536) (0.012,0.571) (0.015,0.571) (0.015,0.679) (0.018,0.679) (0.018,0.714) (0.04,0.714) (0.04,0.786) (0.048,0.786) (0.048,0.821) (0.094,0.821) (0.094,0.857) (0.108,0.857) (0.108,0.893) (0.24,0.893) (0.24,0.929) (0.259,0.929) (0.259,0.964) (0.375,0.964) (0.375,1.0) (0.376,1.0) (0.378,1.0) (1.0,1.0)};
        \addlegendentry{\textbf{AUC}}
        \addlegendentry{\textbf{0.9815}}
        \addlegendentry{0.9530}

    \nextgroupplot[title=Estr. sáfica]
        \addlegendimage{empty legend}
        \addplot[blue] coordinates {(0.0,0.0) (0.0,0.036) (0.0,0.393) (0.002,0.393) (0.002,0.536) (0.003,0.536) (0.003,0.643) (0.006,0.643) (0.006,0.679) (0.009,0.679) (0.009,0.714) (0.011,0.714) (0.011,0.75) (0.014,0.75) (0.014,0.821) (0.028,0.821) (0.028,0.857) (0.041,0.857) (0.041,0.893) (0.052,0.893) (0.052,0.929) (0.065,0.929) (0.065,0.964) (0.077,0.964) (0.077,1.0) (0.088,1.0) (0.09,1.0) (1.0,1.0)};
        \addplot[red] coordinates {(0.0,0.0) (0.0,0.036) (0.0,0.357) (0.001,0.357) (0.001,0.464) (0.002,0.464) (0.002,0.5) (0.005,0.5) (0.005,0.571) (0.008,0.571) (0.008,0.607) (0.01,0.607) (0.01,0.643) (0.012,0.643) (0.012,0.714) (0.013,0.714) (0.015,0.714) (0.029,0.714) (0.029,0.75) (0.036,0.75) (0.036,0.821) (0.052,0.821) (0.052,0.857) (0.117,0.857) (0.117,0.893) (0.201,0.893) (0.201,0.929) (0.216,0.929) (0.216,0.964) (0.487,0.964) (0.487,1.0) (1.0,1.0)};
        \addlegendentry{\textbf{AUC}}
        \addlegendentry{\textbf{0.9881}}
        \addlegendentry{0.9560}

    \nextgroupplot[title=Haiku]
        \addlegendimage{empty legend}
        \addplot[blue] coordinates {(0.0,0.0) (0.0,0.042) (0.0,0.292) (0.001,0.292) (0.001,0.625) (0.002,0.625) (0.002,0.667) (0.003,0.667) (0.003,0.792) (0.004,0.792) (0.004,0.875) (0.005,0.875) (0.005,0.917) (0.006,0.917) (0.006,0.958) (0.008,0.958) (0.008,1.0) (0.607,1.0) (0.609,1.0) (1.0,1.0)};
        \addplot[red] coordinates {(0.0,0.0) (0.0,0.042) (0.0,0.75) (0.002,0.75) (0.002,0.833) (0.003,0.833) (0.003,0.917) (0.006,0.917) (0.006,0.958) (0.02,0.958) (0.02,1.0) (0.469,1.0) (0.471,1.0) (1.0,1.0)};
        \addlegendentry{\textbf{AUC}}
        \addlegendentry{0.9981}
        \addlegendentry{\textbf{0.9985}}

    \nextgroupplot[title=Lira]
        \addlegendimage{empty legend}
        \addplot[blue] coordinates {(0.0,0.0) (0.0,0.036) (0.0,0.214) (0.002,0.214) (0.002,0.357) (0.003,0.357) (0.003,0.393) (0.005,0.393) (0.005,0.607) (0.008,0.607) (0.008,0.643) (0.009,0.643) (0.009,0.679) (0.01,0.679) (0.01,0.714) (0.012,0.714) (0.012,0.786) (0.014,0.786) (0.014,0.821) (0.018,0.821) (0.018,0.857) (0.021,0.857) (0.021,0.893) (0.06,0.893) (0.06,0.929) (0.075,0.929) (0.075,0.964) (0.658,0.964) (0.658,1.0) (0.737,1.0) (0.739,1.0) (1.0,1.0)};
        \addplot[red] coordinates {(0.0,0.0) (0.0,0.036) (0.0,0.071) (0.001,0.071) (0.001,0.214) (0.002,0.214) (0.002,0.357) (0.003,0.357) (0.003,0.464) (0.004,0.464) (0.004,0.536) (0.005,0.536) (0.005,0.571) (0.009,0.571) (0.009,0.607) (0.02,0.607) (0.02,0.714) (0.032,0.714) (0.032,0.75) (0.038,0.75) (0.038,0.786) (0.04,0.786) (0.04,0.821) (0.073,0.821) (0.073,0.857) (0.087,0.857) (0.087,0.893) (0.099,0.893) (0.099,0.929) (0.152,0.929) (0.152,0.964) (0.414,0.964) (0.414,1.0) (0.923,1.0) (0.925,1.0) (1.0,1.0)};
        \addlegendentry{\textbf{AUC}}
        \addlegendentry{\textbf{0.9663}}
        \addlegendentry{0.9629}

    \nextgroupplot[title=Novena]
        \addlegendimage{empty legend}
        \addplot[blue] coordinates {(0.0,0.0) (0.0,0.034) (0.0,0.138) (0.001,0.138) (0.001,0.172) (0.002,0.172) (0.002,0.414) (0.003,0.414) (0.003,0.448) (0.004,0.448) (0.004,0.517) (0.009,0.517) (0.009,0.552) (0.033,0.552) (0.033,0.586) (0.041,0.586) (0.041,0.621) (0.048,0.621) (0.048,0.655) (0.083,0.655) (0.083,0.69) (0.139,0.69) (0.139,0.724) (0.142,0.724) (0.142,0.759) (0.16,0.759) (0.16,0.793) (0.191,0.793) (0.191,0.828) (0.215,0.828) (0.215,0.862) (0.384,0.862) (0.384,0.897) (0.385,0.897) (0.387,0.897) (0.456,0.897) (0.456,0.931) (0.475,0.931) (0.475,0.966) (0.729,0.966) (0.729,1.0) (1.0,1.0)};
        \addplot[red] coordinates {(0.0,0.0) (0.0,0.034) (0.0,0.207) (0.001,0.207) (0.001,0.379) (0.003,0.379) (0.003,0.414) (0.008,0.414) (0.008,0.448) (0.012,0.448) (0.012,0.483) (0.023,0.483) (0.023,0.517) (0.066,0.517) (0.066,0.552) (0.091,0.552) (0.091,0.586) (0.099,0.586) (0.099,0.621) (0.108,0.621) (0.108,0.655) (0.11,0.655) (0.11,0.69) (0.12,0.69) (0.12,0.724) (0.17,0.724) (0.17,0.759) (0.245,0.759) (0.245,0.793) (0.408,0.793) (0.408,0.828) (0.473,0.828) (0.473,0.862) (0.486,0.862) (0.486,0.897) (0.517,0.897) (0.517,0.931) (0.552,0.931) (0.555,0.931) (0.573,0.931) (0.573,0.966) (0.655,0.966) (0.655,1.0) (1.0,1.0)};
        \addlegendentry{\textbf{AUC}}
        \addlegendentry{\textbf{0.8922}}
        \addlegendentry{0.8562}

    \nextgroupplot[title=Octava]
        \addlegendimage{empty legend}
        \addplot[blue] coordinates {(0.0,0.0) (0.0,0.037) (0.0,0.407) (0.002,0.407) (0.002,0.481) (0.003,0.481) (0.003,0.556) (0.004,0.556) (0.004,0.593) (0.005,0.593) (0.005,0.63) (0.007,0.63) (0.007,0.667) (0.015,0.667) (0.015,0.704) (0.021,0.704) (0.021,0.741) (0.028,0.741) (0.028,0.815) (0.039,0.815) (0.039,0.852) (0.052,0.852) (0.052,0.889) (0.069,0.889) (0.069,0.926) (0.07,0.926) (0.07,0.963) (0.074,0.963) (0.074,1.0) (0.908,1.0) (0.91,1.0) (1.0,1.0)};
        \addplot[red] coordinates {(0.0,0.0) (0.0,0.037) (0.0,0.148) (0.001,0.148) (0.001,0.185) (0.002,0.185) (0.002,0.222) (0.003,0.222) (0.003,0.444) (0.004,0.444) (0.004,0.481) (0.009,0.481) (0.009,0.556) (0.012,0.556) (0.012,0.667) (0.014,0.667) (0.014,0.704) (0.018,0.704) (0.018,0.741) (0.029,0.741) (0.029,0.815) (0.03,0.815) (0.03,0.889) (0.123,0.889) (0.123,0.926) (0.239,0.926) (0.239,0.963) (0.408,0.963) (0.408,1.0) (0.813,1.0) (0.815,1.0) (1.0,1.0)};
        \addlegendentry{\textbf{AUC}}
        \addlegendentry{\textbf{0.9844}}
        \addlegendentry{0.9630}

    \nextgroupplot[title=Octava real]
        \addlegendimage{empty legend}
        \addplot[blue] coordinates {(0.0,0.0) (0.001,0.0) (0.001,0.643) (0.002,0.643) (0.002,0.821) (0.003,0.821) (0.003,0.857) (0.004,0.857) (0.004,0.893) (0.005,0.893) (0.005,0.964) (0.008,0.964) (0.008,1.0) (0.449,1.0) (0.451,1.0) (1.0,1.0)};
        \addplot[red] coordinates {(0.0,0.0) (0.0,0.036) (0.0,0.786) (0.001,0.786) (0.001,0.821) (0.002,0.821) (0.002,0.857) (0.005,0.857) (0.005,0.893) (0.069,0.893) (0.069,0.929) (0.071,0.929) (0.071,0.964) (0.681,0.964) (0.681,1.0) (0.683,1.0) (0.686,1.0) (1.0,1.0)};
        \addlegendentry{\textbf{AUC}}
        \addlegendentry{\textbf{0.9981}}
        \addlegendentry{0.9702}

    \nextgroupplot[title=Octavilla]
        \addlegendimage{empty legend}
        \addplot[blue] coordinates {(0.0,0.0) (0.0,0.034) (0.0,0.103) (0.001,0.103) (0.001,0.448) (0.002,0.448) (0.002,0.655) (0.003,0.655) (0.003,0.793) (0.01,0.793) (0.01,0.828) (0.023,0.828) (0.023,0.897) (0.038,0.897) (0.038,0.931) (0.056,0.931) (0.056,1.0) (0.733,1.0) (0.735,1.0) (1.0,1.0)};
        \addplot[red] coordinates {(0.0,0.0) (0.0,0.034) (0.0,0.379) (0.001,0.379) (0.001,0.517) (0.002,0.517) (0.002,0.586) (0.003,0.586) (0.003,0.724) (0.004,0.724) (0.004,0.793) (0.006,0.793) (0.006,0.897) (0.028,0.897) (0.028,0.931) (0.041,0.931) (0.041,0.966) (0.049,0.966) (0.049,1.0) (0.336,1.0) (0.338,1.0) (1.0,1.0)};
        \addlegendentry{\textbf{AUC}}
        \addlegendentry{0.9917}
        \addlegendentry{\textbf{0.9944}}

    \nextgroupplot[title=Ovillejo]
        \addlegendimage{empty legend}
        \addplot[blue] coordinates {(0.0,0.0) (0.0,0.04) (0.0,0.76) (0.001,0.76) (0.001,0.84) (0.002,0.84) (0.002,0.92) (0.003,0.92) (0.003,0.96) (0.049,0.96) (0.049,1.0) (0.972,1.0) (0.974,1.0) (1.0,1.0)};
        \addplot[red] coordinates {(0.0,0.0) (0.0,0.04) (0.0,0.8) (0.001,0.8) (0.001,0.92) (0.189,0.92) (0.189,0.96) (0.311,0.96) (0.311,1.0) (0.943,1.0) (0.945,1.0) (1.0,1.0)};
        \addlegendentry{\textbf{AUC}}
        \addlegendentry{\textbf{0.9977}}
        \addlegendentry{0.9799}
        
%   \end{groupplot}
% \end{tikzpicture}
%     \caption{Caption}
%     \label{fig:my_labelA}
% \end{figure*}

% \begin{figure*}
%     \ContinuedFloat
%     \centering
% \begin{tikzpicture}
%   \begin{groupplot}[
%         clip=false,
%         % axis lines=left, xtick=\empty, ytick=\empty, 
%         % axis lines=none,
%         group style={
%             group size=5 by 4,
%             x descriptions at=edge bottom,
%             y descriptions at=edge left,
%             horizontal sep=10pt,
%             vertical sep=20pt},
%          ymin=0.0,ymax=1.0,
%          xmin=0.0,xmax=1.0,
%          width=\myaxiswidth,
%          height=\myaxiswidth,
%          scale only axis,
%          enlargelimits=0.05,
%          title style={
%             yshift=-4.5pt,
%             % fill=black!10,
%             minimum width=\myaxiswidth},
%          % domain=0:45, %just for example
%          legend style={draw=none},
%          legend pos=south east,
%          samples=20 %just for example
%          ]

    \nextgroupplot[title=Quinteto]
        \addlegendimage{empty legend}
        \addplot[blue] coordinates {(0.0,0.0) (0.0,0.056) (0.0,0.5) (0.002,0.5) (0.002,0.556) (0.004,0.556) (0.004,0.611) (0.011,0.611) (0.011,0.667) (0.02,0.667) (0.02,0.722) (0.021,0.722) (0.021,0.778) (0.023,0.778) (0.023,0.833) (0.037,0.833) (0.037,0.889) (0.113,0.889) (0.113,0.944) (0.116,0.944) (0.116,1.0) (0.409,1.0) (0.411,1.0) (1.0,1.0)};
        \addplot[red] coordinates {(0.0,0.0) (0.0,0.056) (0.0,0.333) (0.001,0.333) (0.001,0.444) (0.007,0.444) (0.007,0.5) (0.009,0.5) (0.009,0.556) (0.011,0.556) (0.011,0.611) (0.014,0.611) (0.014,0.667) (0.069,0.667) (0.069,0.722) (0.1,0.722) (0.1,0.833) (0.102,0.833) (0.102,0.889) (0.111,0.889) (0.111,0.944) (0.149,0.944) (0.149,1.0) (0.64,1.0) (0.642,1.0) (1.0,1.0)};
        \addlegendentry{\textbf{AUC}}
        \addlegendentry{\textbf{0.9807}}
        \addlegendentry{0.9625}

    \nextgroupplot[title=Quintilla]
        \addlegendimage{empty legend}
        \addplot[blue] coordinates {(0.0,0.0) (0.0,0.036) (0.0,0.5) (0.003,0.5) (0.003,0.571) (0.004,0.571) (0.004,0.679) (0.005,0.679) (0.005,0.714) (0.008,0.714) (0.008,0.75) (0.013,0.75) (0.013,0.821) (0.02,0.821) (0.02,0.857) (0.022,0.857) (0.022,0.893) (0.023,0.893) (0.023,0.929) (0.024,0.929) (0.024,0.964) (0.085,0.964) (0.085,1.0) (1.0,1.0)};
        \addplot[red] coordinates {(0.0,0.0) (0.0,0.036) (0.0,0.143) (0.001,0.143) (0.001,0.429) (0.002,0.429) (0.002,0.464) (0.003,0.464) (0.003,0.5) (0.005,0.5) (0.005,0.607) (0.007,0.607) (0.007,0.643) (0.008,0.643) (0.008,0.679) (0.009,0.679) (0.009,0.714) (0.015,0.714) (0.015,0.75) (0.023,0.75) (0.023,0.786) (0.034,0.786) (0.034,0.821) (0.042,0.821) (0.042,0.857) (0.048,0.857) (0.05,0.857) (0.052,0.857) (0.052,0.893) (0.156,0.893) (0.156,0.929) (0.194,0.929) (0.194,0.964) (0.402,0.964) (0.402,1.0) (1.0,1.0)};
        \addlegendentry{\textbf{AUC}}
        \addlegendentry{\textbf{0.9917}}
        \addlegendentry{0.9653}

    \nextgroupplot[title=Redondilla]
        \addlegendimage{empty legend}
        \addplot[blue] coordinates {(0.0,0.0) (0.0,0.034) (0.001,0.034) (0.001,0.172) (0.002,0.172) (0.002,0.31) (0.003,0.31) (0.003,0.345) (0.005,0.345) (0.005,0.379) (0.007,0.379) (0.007,0.414) (0.009,0.414) (0.009,0.448) (0.016,0.448) (0.016,0.483) (0.017,0.483) (0.017,0.517) (0.026,0.517) (0.026,0.586) (0.038,0.586) (0.038,0.621) (0.043,0.621) (0.043,0.655) (0.047,0.655) (0.047,0.69) (0.048,0.69) (0.048,0.759) (0.05,0.759) (0.05,0.862) (0.065,0.862) (0.065,0.897) (0.066,0.897) (0.066,0.931) (0.1,0.931) (0.1,0.966) (0.227,0.966) (0.227,1.0) (0.628,1.0) (0.63,1.0) (1.0,1.0)};
        \addplot[red] coordinates {(0.0,0.0) (0.0,0.034) (0.0,0.069) (0.001,0.069) (0.001,0.207) (0.002,0.207) (0.002,0.241) (0.003,0.241) (0.003,0.276) (0.006,0.276) (0.006,0.345) (0.01,0.345) (0.01,0.379) (0.011,0.379) (0.011,0.414) (0.014,0.414) (0.014,0.448) (0.016,0.448) (0.016,0.483) (0.023,0.483) (0.023,0.517) (0.025,0.517) (0.025,0.586) (0.029,0.586) (0.029,0.621) (0.034,0.621) (0.034,0.655) (0.036,0.655) (0.036,0.69) (0.046,0.69) (0.046,0.724) (0.049,0.724) (0.049,0.759) (0.081,0.759) (0.081,0.793) (0.087,0.793) (0.087,0.828) (0.096,0.828) (0.096,0.862) (0.109,0.862) (0.109,0.897) (0.11,0.897) (0.11,0.931) (0.111,0.931) (0.111,0.966) (0.177,0.966) (0.177,1.0) (0.682,1.0) (0.684,1.0) (1.0,1.0)};
        \addlegendentry{\textbf{AUC}}
        \addlegendentry{\textbf{0.9672}}
        \addlegendentry{0.9618}

    \nextgroupplot[title=Romance]
        \addlegendimage{empty legend}
        \addplot[blue] coordinates {(0.0,0.0) (0.0,0.038) (0.0,0.385) (0.001,0.385) (0.001,0.654) (0.002,0.654) (0.002,0.808) (0.003,0.808) (0.003,0.846) (0.007,0.846) (0.007,0.885) (0.066,0.885) (0.066,0.923) (0.302,0.923) (0.304,0.923) (0.396,0.923) (0.396,0.962) (0.584,0.962) (0.584,1.0) (1.0,1.0)};
        \addplot[red] coordinates {(0.0,0.0) (0.0,0.038) (0.0,0.731) (0.001,0.731) (0.001,0.769) (0.003,0.769) (0.003,0.808) (0.01,0.808) (0.01,0.846) (0.069,0.846) (0.069,0.923) (0.368,0.923) (0.368,0.962) (0.775,0.962) (0.777,0.962) (0.922,0.962) (0.922,1.0) (1.0,1.0)};
        \addlegendentry{\textbf{AUC}}
        \addlegendentry{\textbf{0.9589}}
        \addlegendentry{0.9448}

    \nextgroupplot[title=Rom. arte mayor]
        \addlegendimage{empty legend}
        \addplot[blue] coordinates {(0.0,0.0) (0.0,0.034) (0.0,0.966) (0.01,0.966) (0.01,1.0) (0.273,1.0) (0.275,1.0) (1.0,1.0)};
        \addplot[red] coordinates {(0.0,0.0) (0.0,0.034) (0.0,0.897) (0.002,0.897) (0.002,1.0) (0.925,1.0) (0.927,1.0) (1.0,1.0)};
        \addlegendentry{\textbf{AUC}}
        \addlegendentry{0.9997}
        \addlegendentry{\textbf{0.9998}}

    \nextgroupplot[title=Seguidilla]
        \addlegendimage{empty legend}
        \addplot[blue] coordinates {(0.0,0.0) (0.0,0.038) (0.0,0.077) (0.001,0.077) (0.001,0.192) (0.002,0.192) (0.002,0.231) (0.003,0.231) (0.003,0.308) (0.004,0.308) (0.004,0.385) (0.005,0.385) (0.005,0.462) (0.006,0.462) (0.006,0.654) (0.008,0.654) (0.008,0.769) (0.01,0.769) (0.01,0.808) (0.012,0.808) (0.012,0.846) (0.017,0.846) (0.017,0.923) (0.022,0.923) (0.022,0.962) (0.028,0.962) (0.028,1.0) (0.978,1.0) (0.981,1.0) (1.0,1.0)};
        \addplot[red] coordinates {(0.0,0.0) (0.0,0.038) (0.0,0.308) (0.002,0.308) (0.002,0.346) (0.003,0.346) (0.003,0.385) (0.006,0.385) (0.006,0.538) (0.007,0.538) (0.007,0.654) (0.009,0.654) (0.009,0.731) (0.01,0.731) (0.01,0.769) (0.019,0.769) (0.019,0.808) (0.031,0.808) (0.031,0.846) (0.041,0.846) (0.041,0.923) (0.058,0.923) (0.058,0.962) (0.076,0.962) (0.076,1.0) (0.902,1.0) (0.904,1.0) (1.0,1.0)};
        \addlegendentry{\textbf{AUC}}
        \addlegendentry{\textbf{0.9927}}
        \addlegendentry{0.9868}

    \nextgroupplot[title=Seguid. compuesta]
        \addlegendimage{empty legend}
        \addplot[blue] coordinates {(0.0,0.0) (0.0,0.037) (0.0,0.407) (0.001,0.407) (0.001,0.444) (0.002,0.444) (0.002,0.593) (0.004,0.593) (0.004,0.667) (0.005,0.667) (0.005,0.778) (0.007,0.778) (0.007,0.852) (0.01,0.852) (0.01,0.926) (0.075,0.926) (0.075,0.963) (0.157,0.963) (0.157,1.0) (0.181,1.0) (0.183,1.0) (1.0,1.0)};
        \addplot[red] coordinates {(0.0,0.0) (0.0,0.037) (0.0,0.444) (0.002,0.444) (0.002,0.519) (0.008,0.519) (0.008,0.593) (0.01,0.593) (0.01,0.63) (0.012,0.63) (0.012,0.741) (0.018,0.741) (0.018,0.815) (0.021,0.815) (0.021,0.852) (0.04,0.852) (0.04,0.889) (0.079,0.889) (0.079,0.926) (0.093,0.926) (0.093,0.963) (0.335,0.963) (0.335,1.0) (0.544,1.0) (0.546,1.0) (1.0,1.0)};
        \addlegendentry{\textbf{AUC}}
        \addlegendentry{\textbf{0.9890}}
        \addlegendentry{0.9752}

    \nextgroupplot[title=Seguid. gitana]
        \addlegendimage{empty legend}
        \addplot[blue] coordinates {(0.0,0.0) (0.0,0.053) (0.0,0.684) (0.001,0.684) (0.001,0.789) (0.002,0.789) (0.002,0.842) (0.005,0.842) (0.005,0.947) (0.099,0.947) (0.099,1.0) (0.978,1.0) (0.98,1.0) (1.0,1.0)};
        \addplot[red] coordinates {(0.0,0.0) (0.0,0.053) (0.0,0.789) (0.002,0.789) (0.002,0.842) (0.005,0.842) (0.005,0.895) (0.007,0.895) (0.007,0.947) (0.126,0.947) (0.126,1.0) (0.91,1.0) (0.912,1.0) (1.0,1.0)};
        \addlegendentry{\textbf{AUC}}
        \addlegendentry{\textbf{0.9940}}
        \addlegendentry{0.9926}

    \nextgroupplot[title=Septeto]
        \addlegendimage{empty legend}
        \addplot[blue] coordinates {(0.0,0.0) (0.001,0.0) (0.018,0.0) (0.018,0.2) (0.092,0.2) (0.094,0.2) (0.112,0.2) (0.112,0.4) (0.142,0.4) (0.142,0.6) (0.656,0.6) (0.656,0.8) (0.711,0.8) (0.711,1.0) (1.0,1.0)};
        \addplot[red] coordinates {(0.0,0.0) (0.001,0.0) (0.024,0.0) (0.024,0.4) (0.106,0.4) (0.106,0.6) (0.204,0.6) (0.204,0.8) (0.426,0.8) (0.426,1.0) (0.785,1.0) (0.787,1.0) (1.0,1.0)};
        \addlegendentry{\textbf{AUC}}
        \addlegendentry{0.6722}
        \addlegendentry{\textbf{0.8432}}

    \nextgroupplot[title=Septilla]
        \addlegendimage{empty legend}
        \addplot[blue] coordinates {(0.0,0.0) (0.0,0.143) (0.0,0.571) (0.003,0.571) (0.003,0.714) (0.034,0.714) (0.034,0.857) (0.05,0.857) (0.05,1.0) (0.385,1.0) (0.387,1.0) (1.0,1.0)};
        \addplot[red] coordinates {(0.0,0.0) (0.0,0.143) (0.0,0.429) (0.001,0.429) (0.001,0.857) (0.005,0.857) (0.005,1.0) (0.835,1.0) (0.837,1.0) (1.0,1.0)};
        \addlegendentry{\textbf{AUC}}
        \addlegendentry{0.9876}
        \addlegendentry{\textbf{0.9989}}

    \nextgroupplot[title=Serventesio]
        \addlegendimage{empty legend}
        \addplot[blue] coordinates {(0.0,0.0) (0.0,0.032) (0.0,0.161) (0.001,0.161) (0.001,0.355) (0.002,0.355) (0.002,0.419) (0.003,0.419) (0.003,0.548) (0.007,0.548) (0.007,0.581) (0.008,0.581) (0.008,0.613) (0.01,0.613) (0.01,0.677) (0.013,0.677) (0.013,0.71) (0.014,0.71) (0.014,0.742) (0.02,0.742) (0.02,0.806) (0.024,0.806) (0.024,0.839) (0.036,0.839) (0.036,0.871) (0.038,0.871) (0.038,0.903) (0.049,0.903) (0.049,0.935) (0.075,0.935) (0.077,0.935) (0.108,0.935) (0.108,0.968) (0.239,0.968) (0.239,1.0) (1.0,1.0)};
        \addplot[red] coordinates {(0.0,0.0) (0.0,0.032) (0.0,0.355) (0.001,0.355) (0.001,0.387) (0.002,0.387) (0.002,0.419) (0.005,0.419) (0.005,0.452) (0.006,0.452) (0.006,0.516) (0.008,0.516) (0.008,0.548) (0.01,0.548) (0.01,0.581) (0.011,0.581) (0.011,0.613) (0.018,0.613) (0.018,0.645) (0.027,0.645) (0.027,0.71) (0.033,0.71) (0.033,0.742) (0.034,0.742) (0.034,0.806) (0.037,0.806) (0.037,0.871) (0.052,0.871) (0.054,0.871) (0.054,0.903) (0.065,0.903) (0.065,0.935) (0.072,0.935) (0.072,0.968) (0.197,0.968) (0.197,1.0) (1.0,1.0)};
        \addlegendentry{\textbf{AUC}}
        \addlegendentry{\textbf{0.9801}}
        \addlegendentry{0.9780}

    \nextgroupplot[title=Sexta rima]
        \addlegendimage{empty legend}
        \addplot[blue] coordinates {(0.0,0.0) (0.0,0.111) (0.0,0.222) (0.001,0.222) (0.001,0.333) (0.004,0.333) (0.004,0.444) (0.005,0.444) (0.005,0.778) (0.064,0.778) (0.064,0.889) (0.083,0.889) (0.083,1.0) (0.347,1.0) (0.349,1.0) (1.0,1.0)};
        \addplot[red] coordinates {(0.0,0.0) (0.0,0.111) (0.0,0.444) (0.001,0.444) (0.001,0.556) (0.008,0.556) (0.008,0.667) (0.021,0.667) (0.021,0.778) (0.029,0.778) (0.029,0.889) (0.03,0.889) (0.03,1.0) (0.402,1.0) (0.404,1.0) (1.0,1.0)};
        \addlegendentry{\textbf{AUC}}
        \addlegendentry{0.9815}
        \addlegendentry{\textbf{0.9901}}

    \nextgroupplot[title=Sexteto]
        \addlegendimage{empty legend}
        \addplot[blue] coordinates {(0.0,0.0) (0.0,0.048) (0.0,0.143) (0.002,0.143) (0.002,0.286) (0.003,0.286) (0.003,0.381) (0.004,0.381) (0.004,0.429) (0.009,0.429) (0.009,0.476) (0.01,0.476) (0.01,0.524) (0.015,0.524) (0.015,0.619) (0.018,0.619) (0.018,0.667) (0.02,0.667) (0.02,0.714) (0.021,0.714) (0.021,0.762) (0.049,0.762) (0.049,0.81) (0.051,0.81) (0.051,0.857) (0.11,0.857) (0.11,0.905) (0.117,0.905) (0.117,0.952) (0.137,0.952) (0.137,1.0) (0.193,1.0) (0.195,1.0) (1.0,1.0)};
        \addplot[red] coordinates {(0.0,0.0) (0.0,0.048) (0.002,0.048) (0.002,0.095) (0.004,0.095) (0.004,0.143) (0.008,0.143) (0.008,0.19) (0.009,0.19) (0.009,0.286) (0.01,0.286) (0.01,0.333) (0.011,0.333) (0.011,0.381) (0.013,0.381) (0.013,0.476) (0.015,0.476) (0.015,0.524) (0.018,0.524) (0.018,0.667) (0.026,0.667) (0.026,0.714) (0.033,0.714) (0.033,0.762) (0.044,0.762) (0.044,0.81) (0.053,0.81) (0.053,0.857) (0.058,0.857) (0.058,0.905) (0.093,0.905) (0.093,0.952) (0.286,0.952) (0.286,1.0) (0.517,1.0) (0.519,1.0) (1.0,1.0)};
        \addlegendentry{\textbf{AUC}}
        \addlegendentry{\textbf{0.9720}}
        \addlegendentry{0.9647}

    \nextgroupplot[title=Sexteto lira]
        \addlegendimage{empty legend}
        \addplot[blue] coordinates {(0.0,0.0) (0.0,0.125) (0.001,0.125) (0.001,0.25) (0.004,0.25) (0.004,0.375) (0.006,0.375) (0.006,0.5) (0.015,0.5) (0.015,0.625) (0.025,0.625) (0.025,0.75) (0.047,0.75) (0.047,0.875) (0.075,0.875) (0.075,1.0) (0.84,1.0) (0.842,1.0) (1.0,1.0)};
        \addplot[red] coordinates {(0.0,0.0) (0.001,0.0) (0.001,0.125) (0.002,0.125) (0.002,0.25) (0.006,0.25) (0.006,0.375) (0.007,0.375) (0.007,0.5) (0.018,0.5) (0.018,0.625) (0.027,0.625) (0.027,0.75) (0.126,0.75) (0.126,0.875) (0.47,0.875) (0.472,0.875) (0.537,0.875) (0.537,1.0) (1.0,1.0)};
        \addlegendentry{\textbf{AUC}}
        \addlegendentry{\textbf{0.9784}}
        \addlegendentry{0.9095}

    \nextgroupplot[title=Sextilla]
        \addlegendimage{empty legend}
        \addplot[blue] coordinates {(0.0,0.0) (0.0,0.033) (0.0,0.433) (0.001,0.433) (0.001,0.5) (0.006,0.5) (0.006,0.633) (0.009,0.633) (0.009,0.667) (0.01,0.667) (0.01,0.7) (0.011,0.7) (0.011,0.767) (0.012,0.767) (0.012,0.833) (0.018,0.833) (0.018,0.867) (0.025,0.867) (0.025,0.9) (0.05,0.9) (0.05,0.933) (0.153,0.933) (0.156,0.933) (0.184,0.933) (0.184,0.967) (0.225,0.967) (0.225,1.0) (1.0,1.0)};
        \addplot[red] coordinates {(0.0,0.0) (0.0,0.033) (0.0,0.067) (0.002,0.067) (0.002,0.267) (0.004,0.267) (0.004,0.333) (0.006,0.333) (0.006,0.367) (0.007,0.367) (0.007,0.633) (0.01,0.633) (0.01,0.667) (0.011,0.667) (0.011,0.7) (0.013,0.7) (0.013,0.733) (0.014,0.733) (0.014,0.767) (0.019,0.767) (0.019,0.8) (0.022,0.8) (0.022,0.833) (0.039,0.833) (0.039,0.867) (0.056,0.867) (0.056,0.9) (0.146,0.9) (0.146,0.933) (0.189,0.933) (0.192,0.933) (0.211,0.933) (0.211,0.967) (0.549,0.967) (0.549,1.0) (1.0,1.0)};
        \addlegendentry{\textbf{AUC}}
        \addlegendentry{\textbf{0.9802}}
        \addlegendentry{0.9610}

    \nextgroupplot[title=Silva arromanzada]
        \addlegendimage{empty legend}
        \addplot[blue] coordinates {(0.0,0.0) (0.001,0.0) (0.002,0.0) (0.002,0.094) (0.005,0.094) (0.005,0.156) (0.006,0.156) (0.006,0.188) (0.007,0.188) (0.007,0.219) (0.008,0.219) (0.008,0.25) (0.01,0.25) (0.01,0.281) (0.012,0.281) (0.013,0.312) (0.013,0.344) (0.015,0.344) (0.015,0.375) (0.018,0.375) (0.018,0.438) (0.019,0.438) (0.019,0.469) (0.022,0.469) (0.022,0.5) (0.023,0.5) (0.023,0.531) (0.025,0.531) (0.025,0.594) (0.026,0.594) (0.026,0.656) (0.027,0.656) (0.027,0.688) (0.029,0.688) (0.029,0.719) (0.031,0.719) (0.031,0.75) (0.034,0.75) (0.034,0.781) (0.042,0.781) (0.042,0.812) (0.045,0.812) (0.045,0.844) (0.063,0.844) (0.063,0.875) (0.068,0.875) (0.068,0.906) (0.11,0.906) (0.11,0.938) (0.111,0.938) (0.111,0.969) (0.447,0.969) (0.447,1.0) (1.0,1.0)};
        \addplot[red] coordinates {(0.0,0.0) (0.0,0.031) (0.006,0.031) (0.006,0.062) (0.007,0.062) (0.007,0.094) (0.01,0.094) (0.01,0.125) (0.011,0.125) (0.011,0.188) (0.012,0.188) (0.012,0.312) (0.015,0.312) (0.015,0.344) (0.017,0.375) (0.023,0.375) (0.023,0.406) (0.032,0.406) (0.032,0.438) (0.034,0.438) (0.034,0.5) (0.037,0.5) (0.037,0.531) (0.04,0.531) (0.04,0.688) (0.043,0.688) (0.043,0.75) (0.047,0.75) (0.047,0.781) (0.049,0.781) (0.049,0.812) (0.052,0.812) (0.052,0.844) (0.053,0.844) (0.053,0.875) (0.087,0.875) (0.087,0.906) (0.166,0.906) (0.166,0.938) (0.229,0.938) (0.229,0.969) (0.348,0.969) (0.348,1.0) (1.0,1.0)};
        \addlegendentry{\textbf{AUC}}
        \addlegendentry{\textbf{0.9596}}
        \addlegendentry{0.9500}

    \nextgroupplot[title=Soleá]
        \addlegendimage{empty legend}
        \addplot[blue] coordinates {(0.0,0.0) (0.0,0.031) (0.0,0.5) (0.001,0.5) (0.001,0.656) (0.002,0.656) (0.002,0.75) (0.003,0.75) (0.003,0.812) (0.004,0.812) (0.004,0.844) (0.005,0.844) (0.005,0.875) (0.01,0.875) (0.01,0.938) (0.013,0.938) (0.013,0.969) (0.026,0.969) (0.026,1.0) (0.644,1.0) (0.646,1.0) (1.0,1.0)};
        \addplot[red] coordinates {(0.0,0.0) (0.0,0.031) (0.0,0.75) (0.005,0.75) (0.005,0.812) (0.006,0.812) (0.006,0.844) (0.011,0.844) (0.011,0.875) (0.012,0.875) (0.012,0.906) (0.015,0.906) (0.015,0.938) (0.021,0.938) (0.021,0.969) (0.053,0.969) (0.053,1.0) (0.848,1.0) (0.85,1.0) (1.0,1.0)};
        \addlegendentry{\textbf{AUC}}
        \addlegendentry{\textbf{0.9973}}
        \addlegendentry{0.9960}

    \nextgroupplot[title=Tercetillo]
        \addlegendimage{empty legend}
        \addplot[blue] coordinates {(0.0,0.0) (0.0,0.1) (0.0,0.4) (0.002,0.4) (0.002,0.5) (0.009,0.5) (0.009,0.6) (0.015,0.6) (0.015,0.8) (0.021,0.8) (0.021,1.0) (0.685,1.0) (0.687,1.0) (1.0,1.0)};
        \addplot[red] coordinates {(0.0,0.0) (0.0,0.1) (0.0,0.4) (0.005,0.4) (0.005,0.5) (0.006,0.5) (0.006,0.6) (0.013,0.6) (0.013,0.7) (0.014,0.7) (0.014,0.8) (0.048,0.8) (0.048,0.9) (0.057,0.9) (0.057,1.0) (0.383,1.0) (0.385,1.0) (1.0,1.0)};
        \addlegendentry{\textbf{AUC}}
        \addlegendentry{\textbf{0.9917}}
        \addlegendentry{0.9857}

    \nextgroupplot[title=Terceto]
        \addlegendimage{empty legend}
        \addplot[blue] coordinates {(0.0,0.0) (0.0,0.029) (0.0,0.265) (0.001,0.265) (0.001,0.324) (0.002,0.324) (0.002,0.471) (0.003,0.471) (0.003,0.5) (0.004,0.5) (0.004,0.588) (0.009,0.588) (0.009,0.618) (0.01,0.618) (0.01,0.647) (0.013,0.647) (0.013,0.735) (0.021,0.735) (0.021,0.765) (0.025,0.765) (0.025,0.794) (0.029,0.794) (0.029,0.853) (0.052,0.853) (0.052,0.882) (0.063,0.882) (0.063,0.912) (0.08,0.912) (0.08,0.941) (0.252,0.941) (0.252,0.971) (0.436,0.971) (0.436,1.0) (0.981,1.0) (0.983,1.0) (1.0,1.0)};
        \addplot[red] coordinates {(0.0,0.0) (0.0,0.029) (0.0,0.059) (0.001,0.059) (0.001,0.324) (0.002,0.324) (0.002,0.353) (0.005,0.353) (0.005,0.412) (0.007,0.412) (0.007,0.471) (0.008,0.471) (0.008,0.529) (0.009,0.529) (0.009,0.588) (0.011,0.588) (0.011,0.647) (0.014,0.647) (0.014,0.706) (0.016,0.706) (0.016,0.735) (0.019,0.735) (0.019,0.765) (0.028,0.765) (0.028,0.794) (0.041,0.794) (0.041,0.824) (0.078,0.824) (0.078,0.853) (0.081,0.853) (0.081,0.912) (0.083,0.912) (0.083,0.941) (0.182,0.941) (0.182,0.971) (0.185,0.971) (0.185,1.0) (0.714,1.0) (0.716,1.0) (1.0,1.0)};
        \addlegendentry{\textbf{AUC}}
        \addlegendentry{0.9685}
        \addlegendentry{\textbf{0.9731}}

    \nextgroupplot[title=Terceto monorrimo]
        \addlegendimage{empty legend}
        \addplot[blue] coordinates {(0.0,0.0) (0.001,0.0) (0.002,0.0) (0.002,0.6) (0.029,0.6) (0.029,0.8) (0.137,0.8) (0.139,0.8) (0.217,0.8) (0.217,1.0) (1.0,1.0)};
        \addplot[red] coordinates {(0.0,0.0) (0.001,0.0) (0.001,0.4) (0.011,0.4) (0.011,0.6) (0.034,0.6) (0.034,0.8) (0.042,0.8) (0.044,0.8) (0.18,0.8) (0.18,1.0) (1.0,1.0)};
        \addlegendentry{\textbf{AUC}}
        \addlegendentry{0.9496}
        \addlegendentry{\textbf{0.9546}}

    \nextgroupplot[title=,axis lines=none]
        \addlegendimage{empty legend}

    \nextgroupplot[title=,axis lines=none]
        \addlegendimage{empty legend}

    \nextgroupplot[title=,axis lines=none]
        \addplot[blue] coordinates {(0.0,0.0)};
        \addplot[red] coordinates {(0.0,0.0)};
        \addlegendentry{\textsc{Alberti}}
        \addlegendentry{mBERT}

    \end{groupplot}

    % \node [below=.6cm] at ($(group c4r8.south)$) {$x$-axis label};
   %  \node [left=.7cm,anchor=south,rotate=90] at ($(group c1r1.south west)!0.5!(group c1r2.north west)$) {$x$-axis label};
   % \node [below=.6cm] at ($(group c1r2.south east)!0.5!(group c2r2.south west)$) {$y$-axis label};
\end{tikzpicture}

    \caption{True positive rate (TPR) against false positive rate (FPR) of the receiver operating characteristic (ROC) curves and their corresponding areas (AUC) for the classification of each stanza type versus the rest after fine-tuning \textsc{Alberti} (blue) and mBERT (red). Best AUC score in bold.}
    \label{fig:roc}
\end{figure*}
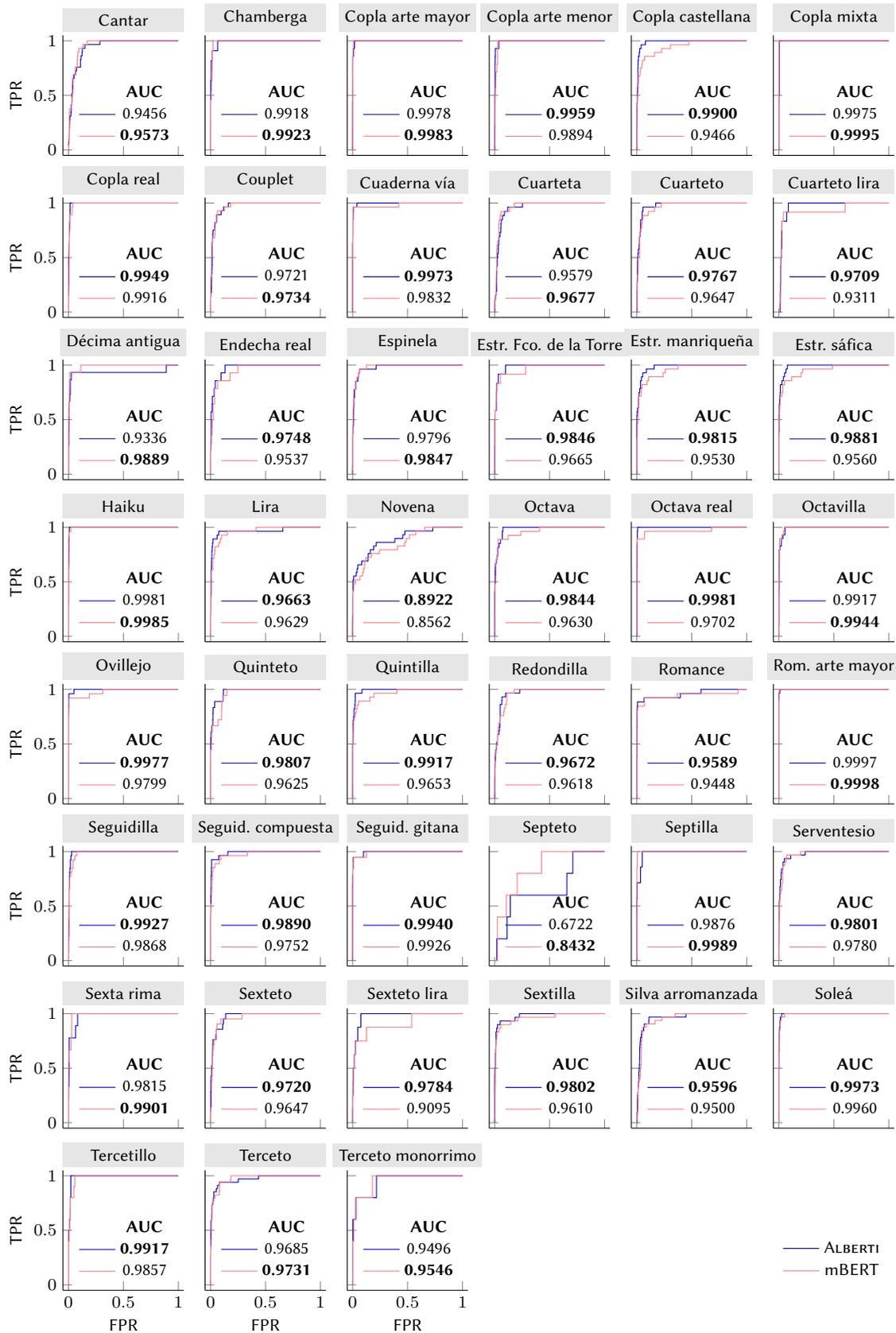

% \begin{table}
% \caption{MLM' accuracy on PULPO validation set. Best scores in bold.}
% \centering
% \begin{tabular}{lrrr}
% \toprule
% Model & English & Spanish & Overall \\
% \midrule
% mBERT & 17.30 & 17.72 & 17.44 \\
% \textsc{Alberti} & \textbf{34.74} & \textbf{36.39} & \textbf{35.27} \\
% \bottomrule
% \end{tabular}
% \label{tab.mlmp}
% \end{table}

\section{Conclusions and Further Work}
% In this work, we have built the first multilingual large language model for poetry using a corpus of verses in 12 languages. We evaluated the model on intrinsic and extrinsic metrics achieving significant improvements over mBERT.
% With the release of the model and the accompanying corpus we hope to foster research in poetry tasks in the fields of Digital Humanities and NLP. We also plan to train \textsc{Alberti} at the stanza level and compare its performance against the current verse-based model.

In this work, we hope to make a significant contribution to the fields of Digital Humanities and NLP by introducing the first multilingual large language model for poetry, \textsc{Alberti}. Our model demonstrated substantial improvements over mBERT, indicating its effectiveness in capturing the nuances of poetic language in various languages and demonstrating the feasibility of domain-specific pre-training for poetry. The evaluation of the model on intrinsic and extrinsic metrics highlights its potential for practical applications in tasks such as stanza-type identification and scansion on a multilingual setting.

The release of our model and accompanying corpus will provide an important resource for researchers in the field, facilitating further investigation into poetry-related tasks. It is our plan to train \textsc{Alberti} at the stanza level and compare its performance against the current verse-based model, which presents itself as an exciting avenue for future research, as it could potentially improve the ability of the model to capture the meaning and structure of poetry in a more sophisticated way. Given the good results obtained by \textsc{Alberti}, despite its training on an arguably outdated model, future iterations will leverage more powerful and larger pre-trained models, thereby enhancing its performance and versatility.

Moreover, we do believe that the strong accuracy of \textsc{Alberti} in the masked language prediction task could pave the way for methods analyzing metaphoric language by leveraging the differences between the predictions of \textsc{Alberti} and the predictions of other models trained on more journalistic or encyclopedic type of data.

Overall, the results of this study have the potential to significantly advance our understanding of poetry in various languages and contribute to the development of more sophisticated NLP models that can capture the subtleties of poetic language. We hope that our work will inspire further research and innovation in this field, and we look forward to seeing how our model and corpus will be used in future studies.

\begin{acknowledgments}
  % Thanks to the developers of ACM consolidated LaTeX styles
  % \url{https://github.com/borisveytsman/acmart} and to the developers
  % of Elsevier updated \LaTeX{} templates
  % \url{https://www.ctan.org/tex-archive/macros/latex/contrib/els-cas-templates}.  
Research for this paper has been partially supported by the Starting Grant research project Poetry Standardization and Linked Open Data: POSTDATA (ERC-2015-STG-679528) obtained by Elena González-Blanco, a project funded by the European Research Council (https://erc.europa.eu) (ERC) under the research and innovation program Horizon2020 of the European Union.

\end{acknowledgments}

%%
%% Define the bibliography file to be used
\bibliography{alberti_sepln2023}

\appendix

\section{PULPO}

PULPO, the Prolific Unannotated Literary Poetry Corpus, is a set of multilingual corpora of verses and stanzas with over 72M words.

\begin{table*}[!htbp]
\caption{Number of poems, verses and words, and the approximate coverage period for each corpus in PULPO.}
\centering
\begin{tabular}{llrrrl}
\toprule
Name & Language &                Poems &               Verses &                Words &                         Period \\
\midrule
THU Chinese Classical Poetry Corpus &   Chinese &              127,682 &              510,728 &            2,553,640 &  $1^{st}$ - $17^{th}$ century \\
TextGrid Poetry Corpus &   German &              105,849 &            3,422,223 &           20,735,344 & $15^{th}$ - $20^{th}$ century \\
SKVR &   Finnish &               89,247 &            1,340,987 &            4,290,341 & $16^{th}$ - $20^{th}$ century \\
Corpus of Czech Verse &   Czech &               66,428 &            2,664,989 &           12,636,867 & $18^{th}$ - $20^{th}$ century \\
Arabic Poetry dataset &   Arabic &               54,300 &               54,944 &            5,328,745 &  $1^{st}$ - $16^{th}$ century \\
Biblioteca Italiana &   Italian &               25,341 &            1,070,717 &            7,121,246 & $15^{th}$ - $20^{th}$ century \\
poesi.as &   Spanish &               25,300 &              910,800 &            5,894,900 & $15^{th}$ - $21^{st}$ century \\
19000 Russian poems &   Russian &               19,315 &              691,361 &            6,559,283 & $15^{th}$ - $20^{th}$ century \\
Poems in Portuguese &   Portuguese &               15,543 &              362,537 &            2,073,420 & $15^{th}$ - $21^{th}$ century \\
ELTE verskorpusz &  Hungarian &                  13,161 &              594,284 &            4,606,974 & $16^{th}$ - $19^{th}$ century \\
Métrique en Ligne &   French &                5,081 &              247,248 &            1,850,222 & $17^{th}$ - $20^{th}$ century \\
Sonetos Siglo de Oro &   Spanish &                5,078 &               65,911 &              466,012 & $16^{th}$ - $17^{th}$ century \\
Disco V3 &   Spanish &                4,530 &               54,066 &              431,428 & $15^{th}$ - $20^{th}$ century \\
Eighteenth Century Poetry Archive &   English &                3,084 &              265,683 &            2,063,668 & $18^{th}$ - $18^{th}$ century \\
German Rhyme Corpus &   German &                1,948 &               47,900 &              270,476 & $17^{th}$ - $20^{th}$ century \\
Stichotheque &   Portuguese &                1,702 &              260,536 &              168,411 & $15^{th}$ - $20^{th}$ century \\
Gongocorpus &   Spanish &                  481 &               20,621 &               99,490 & $16^{th}$ - $17^{th}$ century \\
Poesía Lírica Castellana Siglo de Oro &   Spanish &                  475 &               51,219 &              299,402 & $16^{th}$ - $17^{th}$ century \\
For Better For Verse &   English &                  103 &                1,084 &               41,749 & $15^{th}$ - $20^{th}$ century \\
A Gutenberg Poetry Corpus &   English &                  N/A\footnotemark &            3,085,117 &           22,124,040 & $15^{th}$ - $20^{th}$ century \\
\bottomrule
\end{tabular}
\label{tab.corpora}
\end{table*}
\footnotetext{The poems as such are not available as lines that "looked like" poetry where extracted from books in the Project Gutenberg. See \url{https://github.com/aparrish/gutenberg-poetry-corpus}.}

The individual corpora were downloaded using the \href{https://github.com/linhd-postdata/averell/}{Averell} tool, developed by the \href{https://postdata.linhd.uned.es/}{POSTDATA} team, and other sources found on the Internet.

\subsection{Averell sources}
\subsubsection{Spanish}
\begin{itemize}
\item \href{https://github.com/pruizf/disco}{Disco v3}
\item \href{https://github.com/bncolorado/CorpusSonetosSigloDeOro}{Corpus of Spanish Golden-Age Sonnets}
\item \href{https://github.com/bncolorado/CorpusGeneralPoesiaLiricaCastellanaDelSigloDeOro}{Corpus general de poesía lírica castellana del Siglo de Oro}
\item \href{https://github.com/linhd-postdata/gongocorpus}{Gongocorpus} - \href{http://obvil.sorbonne-universite.site/corpus/gongora/gongora_obra-poetica}{source}
\end{itemize}

\subsubsection{English}
\begin{itemize}
\item \href{https://github.com/alhuber1502/ECPA}{Eighteenth-Century Poetry Archive (ECPA)}
\item \href{https://github.com/waynegraham/for_better_for_verse}{For better for verse}
\end{itemize}

\subsubsection{French}
\begin{itemize}
\item \href{https://crisco2.unicaen.fr/verlaine/index.php?navigation=accueil}{Métrique en Ligne} - \href{https://github.com/linhd-postdata/metrique-en-ligne}{source}
\end{itemize}

\subsubsection{Italian}
\begin{itemize}
\item \href{https://github.com/linhd-postdata/biblioteca_italiana}{Biblioteca italiana} - \href{http://www.bibliotecaitaliana.it/}{source}
\end{itemize}

\subsubsection{Czech}
\begin{itemize}
\item \href{https://github.com/versotym/corpusCzechVerse}{Corpus of Czech Verse}
\end{itemize}

\subsubsection{Portuguese}
\begin{itemize}
\item \href{https://gitlab.com/stichotheque/stichotheque-pt}{Stichotheque}
\end{itemize}

% We additionally obtained the following corpora from these sources.

\subsection{Internet sources}
\subsubsection{Spanish}
\begin{itemize}
\item \href{https://github.com/linhd-postdata/poesi.as}{Poesi.as} - \href{http://www.poesi.as/}{source}
\end{itemize}

\subsubsection{English}
\begin{itemize}
\item \href{https://github.com/aparrish/gutenberg-poetry-corpus}{A Gutenberg Poetry Corpus}
\end{itemize}

\subsubsection{Arabic}
\begin{itemize}
\item \href{https://www.kaggle.com/ahmedabelal/arabic-poetry}{Arabic Poetry dataset}
\end{itemize}

\subsubsection{Chinese}
\begin{itemize}
\item \href{https://github.com/THUNLP-AIPoet/Datasets/tree/master/CCPC}{THU Chinese Classical Poetry Corpus}
\end{itemize}

\subsubsection{Finnish}
\begin{itemize}
\item \href{https://github.com/sks190/SKVR}{SKVR}
\end{itemize}

\subsubsection{German}
\begin{itemize}
\item \href{https://github.com/linhd-postdata/textgrid-poetry}{TextGrid Poetry Corpus} - \href{https://textgrid.de/en/digitale-bibliothek}{source}
\item \href{https://github.com/tnhaider/german-rhyme-corpus}{German Rhyme Corpus}
\end{itemize}

\subsubsection{Hungarian}
\begin{itemize}
\item \href{https://github.com/ELTE-DH/verskorpusz}{ELTE verskorpusz}
\end{itemize}

\subsubsection{Portuguese}
\begin{itemize}
\item \href{https://www.kaggle.com/oliveirasp6/poems-in-portuguese}{Poems in Portuguese}
\end{itemize}

\subsubsection{Russian}
\begin{itemize}
\item \href{https://www.kaggle.com/grafstor/19-000-russian-poems}{19,000 Russian poems}
\end{itemize}

\section{Availability}
\begin{itemize}
\item \textsc{Alberti}: \url{https://huggingface.co/linhd-postdata/alberti}
\item \textsc{PULPO}: \url{https://huggingface.co/datasets/linhd-postdata/pulpo}
\end{itemize}

\end{document}